\documentclass[runningheads]{llncs}

\usepackage{eccv}

\usepackage{eccvabbrv}

\usepackage{graphicx}
\usepackage{booktabs}
\usepackage{url}
\usepackage[accsupp]{axessibility}  % Improves PDF readability for those with disabilities.

\usepackage{hyperref}

\usepackage{orcidlink}

\newcommand{\methodName}{ROAR-3D\xspace}

\begin{document}

\title{\textbf{\methodName}: ROuting ARbitrary Views for High-Fidelity 3D Generation}

\titlerunning{ROAR-3D}

\author{
  Hanxiao Sun\inst{1,2}\textsuperscript{*,$\diamond$}\and
  Mingxin Yang\inst{2}\textsuperscript{*} \and
  Shuhui Yang\inst{2} \and
  Zebin He\inst{1,2} \and
  Xintong Han\inst{2} \and
  Hongbo Fu\inst{1} \and
  Chunchao Guo\inst{2}\textsuperscript{\dag} \and
  Wenhan Luo\inst{1}\textsuperscript{\dag}
}

% Second Author\inst{2,3}\orcidlink{1111-2222-3333-4444} \and
% Third Author\inst{3}\orcidlink{2222--3333-4444-5555}}

\authorrunning{Sun et al.}

\institute{The Hong Kong University of Science and Technology \and
Tencent Hunyuan
}

% Springer Heidelberg, Tiergartenstr.~17, 69121 Heidelberg, Germany

\maketitle
\renewcommand{\thefootnote}{}
\footnotetext{$\diamond$ Work done during internship at Tencent Hunyuan.}
\footnotetext{* Equal contribution.}
\footnotetext{$\dag$ Corresponding author.}
% ---- Teaser figure ----
\begin{figure}[h]
\begin{center}
  \vspace{-4mm}
  \includegraphics[width=\linewidth]{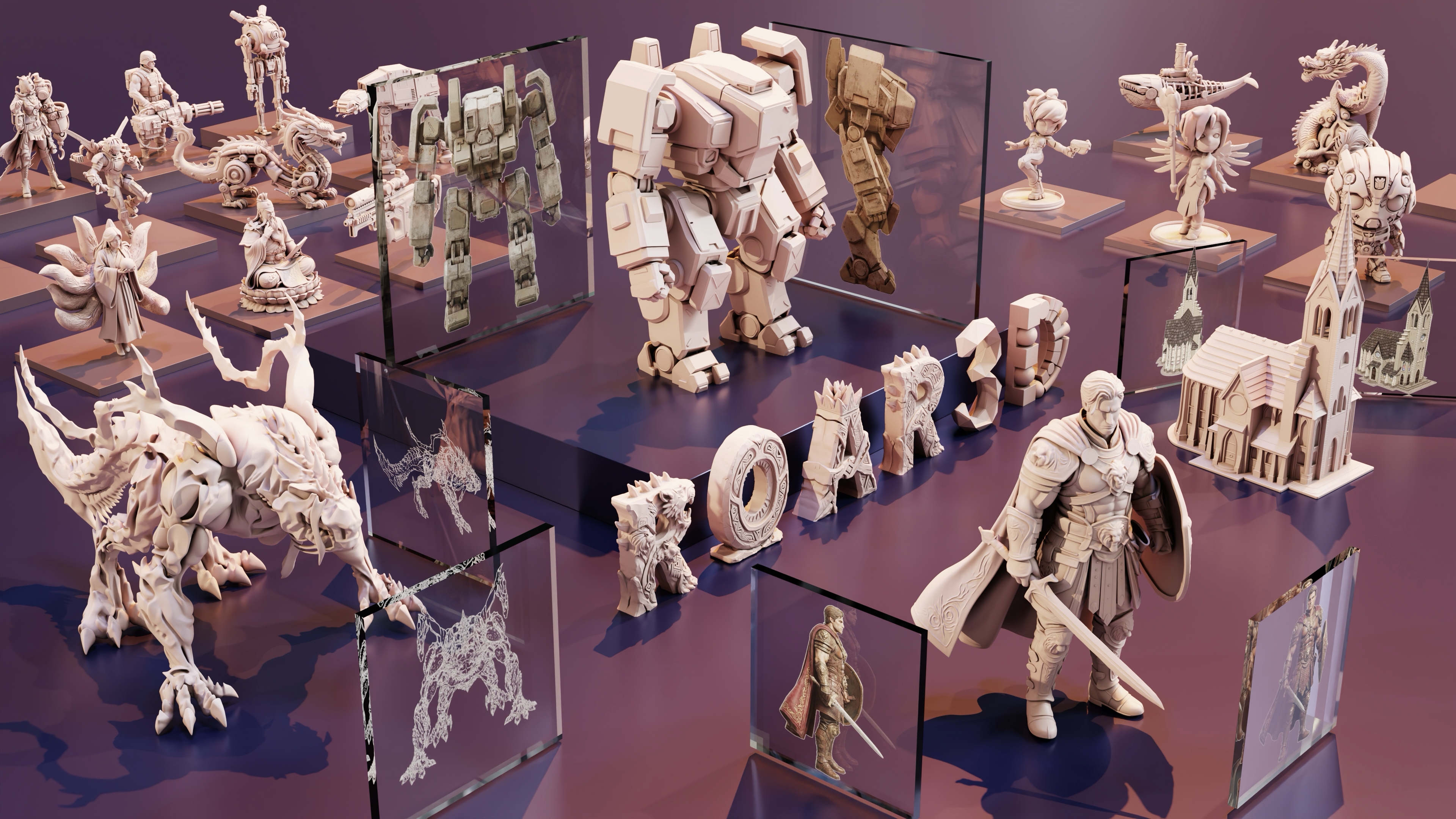}
  \vspace{-8mm}
\end{center}
  \caption{Professional-grade 3D assets generated by \textbf{ROAR-3D} from arbitrary collections of concept art, design sketches, and video frames, exhibiting hand-crafted fidelity.}
\label{fig:teaser}
\end{figure}

\begin{abstract}
Single-image-to-3D generative models can now produce high-quality geometry, yet conditioning on a single view inevitably introduces ambiguity about unseen regions. Multi-view conditioning can reduce this ambiguity, but existing methods either require fixed canonical viewpoints or rely on external reconstruction modules that impose heavy training costs and limit generation quality. We observe that pretrained single-view models already possess strong 2D-to-3D grounding that can be reused for multi-view conditioning. However, a closer analysis reveals that their conditioning mechanism entangles orientation control with geometry transfer---two functions that conflict when images from different viewpoints are naively combined. Based on this analysis, we propose \textbf{ROAR-3D}, a lightweight method that upgrades a pretrained single-view model to accept an arbitrary number of unposed images. A \emph{token-wise view router} assigns each 3D latent token to its most relevant view, implicitly establishing 2D-to-3D correspondences without explicit pose input. A \emph{dual-stream attention} design preserves the pretrained primary-view behavior while routing auxiliary views through a separate path dedicated to geometric enrichment. An \emph{orientation perturbation strategy} ensures the auxiliary path learns orientation-independent geometry transfer. These components introduce minimal trainable parameters and add zero inference overhead relative to the single-view baseline. ROAR-3D achieves state-of-the-art multi-view 3D generation quality and supports test-time view scaling from 1 to 12+ views with consistent improvements.
\end{abstract}

\section{Introduction}
\label{sec:intro}

Recent advances in diffusion-based 3D generative models have enabled remarkably high-quality 3D geometry generation from a single image~\cite{li2024craftsman, zhang2024clay, xiang2024trellis, zhao2025hunyuan3d2, li2025triposg, ye2025hi3dgen, chen2025ultra3d, lai2025hunyuan3d25, lai2025lattice, xiang2025trellis2}, producing results that increasingly approach the quality of handcrafted 3D assets. By leveraging generative priors learned from large-scale 3D datasets, these methods can synthesize complete and plausible geometry even for regions not visible in the input. However, conditioning on a single view inevitably introduces ambiguity: the model must hallucinate unseen structures, and the generated output may not faithfully reflect the user's intended geometry from other viewpoints. This limits the practical utility of single-view methods in applications where the generated 3D geometry must closely match the appearance specified by multiple reference images.

A natural solution is to condition the generation on multiple views. Multi-view conditioned generation occupies a unique position between pure generation and multi-view reconstruction: it retains the ability to complete unobserved regions and tolerate inconsistencies across input views (as it samples from a learned 
conditional distribution)---unlike reconstruction methods~\cite{mildenhall2020nerf, kerbl20233dgs, wang2025vggt} that struggle with occlusion, incomplete coverage, occlusion, weak texture and require strictly consistent multi-view inputs---while significantly reducing ambiguity by constraining the output to be consistent with multiple observed images. This generative nature also opens up workflows that are inaccessible to traditional reconstruction, such as producing 3D assets directly from concept art, design sketches, or video frames (see Fig.~\ref{fig:teaser}), substantially reducing the manual effort required in professional 3D content creation pipelines. 

Despite their potential, existing multi-view conditioned 3D generation methods face notable limitations, which can be broadly categorized into two categories. The first category adopts fixed viewpoint configurations, requiring input images to be captured from a predefined set of canonical poses. This includes feed-forward reconstruction-generation hybrids~\cite{honglrm, tang2025lgm, xu2024instantmesh} as well as commercial systems~\cite{hunyuan3d, tripo3d, hitem3d} that only support a small number of predetermined viewpoints, severely limiting their applicability to unconstrained real-world inputs. The second attempts to incorporate flexible multi-view information by integrating external reconstruction modules---such as VGGT~\cite{wang2025vggt} features---into the generative pipeline~\cite{chang2025reconviagenaccuratemultiview3d}. However, the domain gap between reconstruction features and the latent generative space imposes a substantial training overhead. Furthermore, The generation fidelity is strictly upper-bounded by the accuracy of the reconstruction module; consequently, when external features are inaccurate or incomplete, the generative model lacks the inherent robustness to rectify these artifacts or plausibly complete the missing regions. A truly flexible system should accept an arbitrary number of images from arbitrary viewpoints without requiring camera poses or external reconstruction as input.

We observe that pretrained single-view image-to-3D models already contain the necessary knowledge for multi-view conditioning. Since these models can generate plausible 3D geometry from an image at any viewpoint, the attention layers that bridge 2D image tokens and 3D latent tokens have, in effect, learned 2D--3D correspondences across different viewing angles. These layers are among the most expensive components to train, as they must establish correspondences across two very different modalities. Rather than re-learning these correspondences, we aim to maximally reuse them. However, naively feeding multiple views into the existing attention layers leads to degraded results. A closer analysis reveals that the conditioning process entangles two distinct roles: (i) defining the global orientation of the output, and (ii) transferring local geometric structure. In the single-view case this coupling is harmless, but with multiple views, conflicting orientation signals from different viewpoints destabilize the generation. The challenge therefore reduces to: \emph{how to let multiple views contribute complementary geometry without disrupting the learned weights or introducing orientation conflicts?}

We introduce \textbf{ROAR-3D}, a lightweight method that upgrades a pretrained single-image-to-3D model to accept arbitrary multi-view input without retraining the backbone or introducing external encoders. At its core is a \emph{token-wise view router} that, for each 3D latent token, selects the single most informative view before the attention computation. Because each 3D latent token carries spatial meaning, this selection naturally establishes 2D-to-3D correspondences across views, sidestepping the need for explicit pose conditioning. The router's projections are initialized from the pretrained attention weights, greatly reducing training cost. To separate orientation control from geometric structure transfer, we adopt a \emph{dual-stream attention} design: the \emph{primary stream} preserves the pretrained behavior for the reference view (setting both orientation and base geometry), while the \emph{auxiliary stream} handles supplementary views (enhancing geometric structure without altering orientation). An \emph{orientation perturbation strategy} further strengthens the auxiliary stream by training it on deliberately misaligned 3D latents, ensuring it learns orientation-independent geometric transfer. Together, these components introduce minimal additional parameters and maintain the same inference cost as the single-view baseline, while enabling the model to dynamically fuse information from any number of unposed views.

Our contributions are as follows:
\begin{itemize}
    \item We propose \textbf{ROAR-3D}, a method that extends pretrained single-view 3D generative models to support multi-view input without requiring external encoders or camera poses.
    \item We introduce a token-wise view router that uses 3D latent tokens as anchors to ground multi-view 2D information in 3D space, combined with a dual-stream attention design and orientation perturbation strategy that explicitly decouple orientation control from geometric structure transfer.
    \item ROAR-3D achieves state-of-the-art multi-view 3D generation quality while maintaining the same inference cost as the single-view baseline, and supports test-time view scaling from 1 to 12+ views with consistent improvements.
\end{itemize}

\section{Related Work}
\label{sec:related}
\begin{figure*}[tb] \centering
    \includegraphics[width=\textwidth]{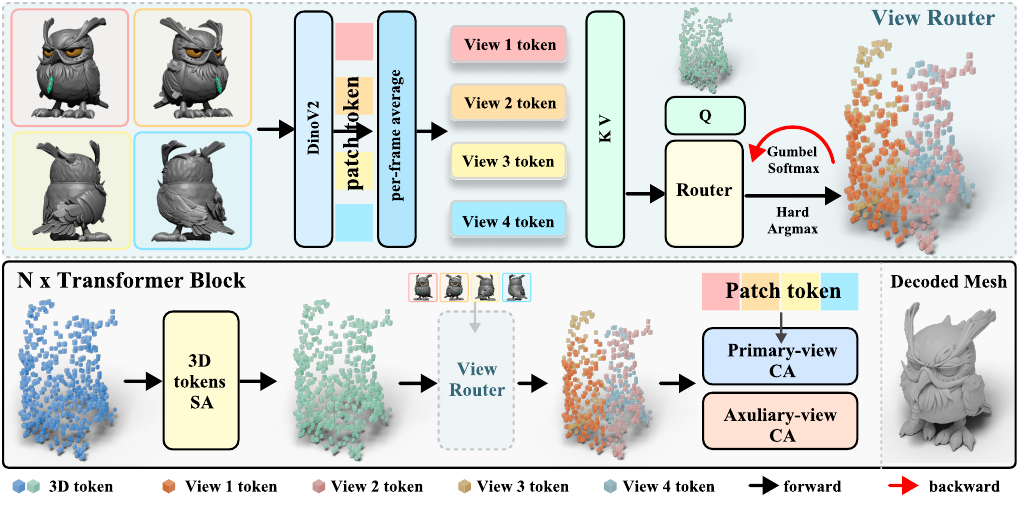}
    \vspace{-6mm}
    \caption{An overview illustration of the proposed \textbf{\methodName} framework, which seamlessly integrates supplementary visual cues from arbitrary unposed views with pre-trained single-view generative priors via a lightweight token-wise routing mechanism for high-fidelity 3D reconstruction.}
    \label{fig:overview}
    \vspace{-3mm}
\end{figure*}

\subsubsection{Single-View 3D Generation.}
Single-view 3D generation aims to lift a single 2D image into a complete 3D representation. Early approaches distill 3D knowledge from pretrained 2D diffusion models via score distillation sampling~\cite{poole2022dreamfusion, dreamgaussian, qiu2024richdreamer, wang2024prolificdreamer, lin2023magic3d, tang2023make}, while another line of work generates multi-view images from a single input and fuses them into 3D representations~\cite{li2023instant3d, xu2024instantmesh, tang2025lgm, xu2024grm, wang2025crm, wei2024meshlrm, liu2024one2345, liu2024one2345pp, xudmv3d, li2024era3d, zuo2024videomv, wu2024unique3d}. To achieve high-fidelity appearance, several approaches specifically focus on texture generation and material decomposition, leveraging generative priors to map 2D appearance onto 3D manifolds~\cite{chen2023text2tex, liu2024text, feng2025romantex, he2025materialmvp,luo2025matpedia}. More recently, 3D native generative methods apply diffusion processes directly on 3D representations such as point clouds~\cite{luo2021diffusion, zhou20213d, nichol2022pointe}, voxel grids~\cite{hui2022neural, tang2023volumediffusion, muller2023diffrf}, triplanes~\cite{chen2023single, wang2023rodin, shue2023triplanediffusion}, and 3D Gaussians~\cite{zhang2024gaussiancube}. Latent 3D diffusion models~\cite{zhang20233dshape2vecset, zhao2024michelangelo, li2024craftsman, zhang2024clay, wu2024direct3d, li2025triposg, zhao2025hunyuan3d2, ye2025hi3dgen} further improve quality by learning mappings between 2D images and compressed 3D latent spaces. These models have established strong 2D-to-3D grounding through large-scale training; however, they are inherently limited by single-view ambiguity and must hallucinate all unseen geometry. Our work builds directly on this family of models, aiming to activate their learned grounding capability for multi-view conditioning.

\subsubsection{Multi-View 3D Reconstruction.}
Multi-view reconstruction recovers 3D geometry from multiple calibrated or uncalibrated images. Classical multi-view stereo (MVS) methods triangulate correspondences across views~\cite{furukawa2015multi, galliani2015massively, schonberger2016pixelwise, xu2019multi}, while learning-based approaches~\cite{yao2018mvsnet, yao2019recurrent, chen2019point, cheng2020deep, gu2020cascade, yang2020cost, wang2021patchmatchnet} improve efficiency via deep networks. NeRF-based methods~\cite{mildenhall2020nerf, lin2021barf, wang2021nerf} and 3D Gaussian Splatting~\cite{kerbl20233dgs} enable high-quality novel view synthesis but typically require accurate camera poses. Recent works such as DUSt3R~\cite{wang2024dust3r} and VGGT~\cite{wang2025vggt} estimate point clouds and poses directly from unposed views. Large Reconstruction Models (LRMs)~\cite{honglrm, wei2024meshlrm, xu2024instantmesh, tang2025lgm, xu2024grm} regress compact 3D representations from multi-view inputs in a feed-forward manner. Despite these advances, reconstruction methods fundamentally require consistent multi-view observations and struggle with occlusion, weak texture, and incomplete coverage. Without generative priors, they tend to produce incomplete geometry or over-smoothed surfaces in unobserved regions.

\subsubsection{Multi-View Conditioned 3D Generation.}
Multi-view conditioned 3D generation combines the completeness of generative models with the accuracy afforded by multiple observations. Existing methods fall into two categories. The first adopts fixed viewpoint configurations: feed-forward models~\cite{honglrm, tang2025lgm, xu2024instantmesh} and commercial systems~\cite{hyper3d, tripo3d, hitem3d, hy3d} accept images only from a small set of predetermined canonical poses, limiting their applicability to unconstrained inputs. The second category integrates external reconstruction modules into the generative pipeline. For instance, \cite{chang2025reconviagenaccuratemultiview3d} injects VGGT~\cite{wang2025vggt} features into a diffusion model to provide reconstruction priors. While effective, the domain gap between reconstruction features and the latent generative space imposes a heavy training burden, and generation quality is bounded by the accuracy of the external module. Other works explore optimization-based pose alignment~\cite{wu2023ifusion} or volume diffusion~\cite{liu2024one2345pp, xu2024sparp}, but face trade-offs between computational cost and output quality. In contrast, ROAR-3D requires no external reconstruction module, no camera poses, and no fixed viewpoint assumption. By introducing a lightweight routing mechanism and dual-stream attention into the pretrained backbone, it enables direct and efficient conditioning on arbitrary unposed multi-view images.
\section{Method}

% hxs
% We propose \methodName, a novel framework that enables arbitrary multi-view conditioning by leveraging the flow-matching Diffusion Transformer (DiT) and 3DShape2VecSet-based VAE of Hunyuan3D 2.1~\cite{hunyuan3d2025hunyuan3d21imageshighfidelity}. Instead of retraining this heavy backbone, we inject supplementary visual information via a lightweight adaptation that preserves the robust priors of the pre-trained model. Specifically, we introduce three key components: (1) a \textbf{token-wise routing mechanism} that enables each 3D latent token to dynamically select the most relevant view based on geometric affinity before cross-attention computation (\S\ref{sec:router}); (2) a \textbf{dual-stream cross-attention design} that processes the reference view via a dedicated primary branch and routed supplementary views via a parallel auxiliary branch (\S\ref{sec:dual_ca}); and (3) an \textbf{alignment decorrelation strategy} that decouples auxiliary views from the canonical geometric alignment, enforcing content-aware routing by eliminating trivial positional correspondences during training (\S\ref{sec:perturbation}).

We present \textbf{ROAR-3D}, a lightweight and effective method that extends a pre-trained single-image-to-3D generative model to accept an arbitrary number of multi-view images as input, so that the generated 3D geometry faithfully captures the structural details specified by every input view. Illustrated in Fig. \ref{fig:overview}, we build upon the flow-matching Diffusion Transformer (DiT) and the 3DShape2VecSet-based VAE of Hunyuan3D 2.1~\cite{hunyuan3d2025hunyuan3d21imageshighfidelity}. To incorporate multi-view information without retraining the backbone from scratch, we introduce three key components: 
a \textbf{token-wise view routing mechanism} (\S\ref{sec:router}) that assigns each 3D latent token to its most informative view by repurposing the pre-trained cross-attention module; 
a \textbf{dual-stream cross-attention design} (\S\ref{sec:dual_ca}) where the primary branch uses a reference view to set 3D orientation and establish its corresponding geometric structure, while the auxiliary branch adds geometric details from other views without changing geometry orientation; 
% and an \textbf{orientation perturbation strategy} (\S\ref{sec:perturbation}) that trains the auxiliary branch to better transfer image geometry to 3D across different orientations, while keeping orientation control in the primary branch. 
and an \textbf{orientation perturbation strategy} (\S\ref{sec:perturbation}) that forces the auxiliary branch to learn an orientation-invariant image-to-3D geometric transfer, while keeping orientation control strictly in the primary branch. 
We further adopt a second-stage refinement model based on LATTICE~\cite{lai2025lattice} with the same multi-view conditioning architecture to enhance the quality of the generated geometry (\S\ref{sec:refinement}).

\subsection{Token-Wise View Router}
\label{sec:router}

In modern image-to-3D DiT models~\cite{lai2025lattice,xiang2025trellis2, chen2025ultra3d, hunyuan3d2025hunyuan3d21imageshighfidelity, zhao2025hunyuan3d2, xiang2024trellis}, the attention layers that bridge 2D image tokens and 3D latent tokens must learn correspondences across different modalities, making them one of the most challenging components to train. Since a well-trained single-view model can already produce plausible 3D geometry from an image at any viewpoint, we argue that it has, in effect, learned 2D--3D correspondences across different viewing angles. The challenge for multi-view extension is therefore not about re-learning these correspondences, but about designing a conditioning mechanism that lets multiple views contribute without disrupting the learned weights.

A na\"ive approach---concatenating all view tokens and feeding them into a single attention layer---introduces two problems: (i) each 3D token must attend to tokens from all views simultaneously, creating ambiguity from conflicting visual signals across views, and (ii) computational cost grows linearly with the number of views. Alternatively, encoding view identity---whether through fixed viewpoint bins with learnable embeddings or through estimated camera parameters---either restricts the model to a predefined set of poses or introduces errors from inaccurate pose estimation. Our router takes a different path: each 3D latent token directly selects its most informative view based on feature affinity, effectively using the 3D tokens themselves as anchors that ground multi-view 2D information in 3D space.

To be specific, the router is applied independently at each transformer block. For each 3D latent token, it computes a routing score against every input view and selects the top-scoring one \emph{before} cross-attention, keeping the per-token attention cost identical to the single-view baseline. Since the router's task---determining which view best matches a given 3D token---is closely related to what the pretrained cross-attention already computes at the token level, we initialize the router's query and key projections directly from the pretrained cross-attention weights, which greatly reduces the training cost of the router. We now describe its architecture in detail.

\subsubsection{Router architecture.}
As illustrated in Fig.~\ref{fig:overview}, given a set of $V$ input views whose DINOv2~\cite{oquab2024dinov2} patch features are $\{\mathbf{F}_v \in \mathbb{R}^{S \times D}\}_{v=1}^{V}$ and 3D latent tokens $\mathbf{Z} \in \mathbb{R}^{N \times D}$, the router at each transformer block~$\ell$ maintains query and key projection matrices $\mathbf{W}_q^{(\ell)}, \mathbf{W}_k^{(\ell)}$ initialized as described above, along with the original QK-normalization (RMSNorm as in~\cite{hunyuan3d2025hunyuan3d21imageshighfidelity}). The router query is derived from each 3D latent token. For the router key, since the router selects among views rather than individual patches, we need a single view-level representation for each input view. A natural candidate is the CLS token, but it encodes a global image summary that does not capture how individual patches relate to 3D tokens. Instead, we use the mean-pooled DINOv2 patch tokens: views with more patches correlated to a given 3D token produce a mean representation naturally closer to it in feature space, yielding more accurate routing:
Let $\bar{\mathbf{f}}_v = \frac{1}{S}\sum_{s=1}^{S} \mathbf{f}_{v,s}$ denote the mean-pooled patch representation of view~$v$, and $\tilde{\mathbf{z}}_i^{(\ell)} = \mathrm{LN}(\mathbf{z}_i^{(\ell)})$ the layer-normalized 3D latent token. The router query and key are:
\begin{equation}
\mathbf{q}_i^{(\ell)} = \mathrm{RMSNorm}\!\left(\mathbf{W}_q^{(\ell)} \, \tilde{\mathbf{z}}_i^{(\ell)}\right), \quad \mathbf{k}_v^{(\ell)} = \mathrm{RMSNorm}\!\left(\mathbf{W}_k^{(\ell)} \, \bar{\mathbf{f}}_v\right),
\end{equation}
where $\mathbf{W}_q^{(\ell)} \in \mathbb{R}^{Hd \times D}$ and $\mathbf{W}_k^{(\ell)} \in \mathbb{R}^{Hd \times D}$ project the inputs into a lower-dimensional space shared with the pretrained cross-attention. Both the pre-projection LayerNorm and post-projection RMSNorm are inherited from the pretrained cross-attention of block~$\ell$.
\subsubsection{Routing logits.}
\label{sec:routing_logits}
Both $\mathbf{q}_i^{(\ell)}$ and $\mathbf{k}_v^{(\ell)}$ are split into $H$ heads of dimension $d$, yielding $\mathbf{q}_{i,h}^{(\ell)}, \mathbf{k}_{v,h}^{(\ell)} \in \mathbb{R}^{d}$. The routing logit is:
\begin{equation}
r_{i,v}^{(\ell)} = \mathbf{w}_{\mathrm{agg}}^\top \left[ \frac{(\mathbf{q}_{i,1}^{(\ell)})^\top \mathbf{k}_{v,1}^{(\ell)}}{\sqrt{d}},\; \ldots,\; \frac{(\mathbf{q}_{i,H}^{(\ell)})^\top \mathbf{k}_{v,H}^{(\ell)}}{\sqrt{d}} \right],
\end{equation}
where $\mathbf{w}_{\mathrm{agg}} \in \mathbb{R}^{H}$ is learnable, initialized to $1/H$. 

\subsubsection{Discrete selection via Gumbel-Softmax.}
\label{sec:gumbel}
Each token must select exactly one view, but $\arg\max$ is not differentiable. We use hard Gumbel-Softmax~\cite{jang2016categorical} with straight-through estimation. During training, Gumbel noise is added to encourage exploration:
\begin{equation}
v_i^{*} = \arg\max_{v} \left( r_{i,v}^{(\ell)} + g_v \right), \quad g_v = -\log(-\log(u_v)),\; u_v \sim \mathrm{Uniform}(0,1).
\end{equation}
Gradient flow is enabled via:
\begin{equation}
\mathbf{y}_i = \mathbf{y}_i^{\mathrm{hard}} - \mathrm{sg}(\mathbf{y}_i^{\mathrm{soft}}) + \mathbf{y}_i^{\mathrm{soft}},
\end{equation}
where $\mathbf{y}_i^{\mathrm{hard}} \in \{0,1\}^V$ is the one-hot selection at $v_i^*$, $\mathbf{y}_i^{\mathrm{soft}} = \mathrm{softmax}((\mathbf{r}_i^{(\ell)} + \mathbf{g})/\tau)$, and $\mathrm{sg}(\cdot)$ is stop-gradient. The forward pass uses the discrete $\mathbf{y}_i^{\mathrm{hard}}$; gradients in the backward pass flow through $\mathbf{y}_i^{\mathrm{soft}}$. At inference, noise is removed and selection is deterministic.

The selected view $v_i^*$ and soft weights $\mathbf{y}_i$ are then passed to the dual-stream cross-attention (\S\ref{sec:dual_ca}).

\subsection{Dual-Stream Cross-Attention}
\label{sec:dual_ca}

In single-view image-to-3D generation, the attention layers that bridge 2D image tokens and 3D latent tokens implicitly serve a dual role: they align the output 3D geometry to a canonical orientation determined by the input viewpoint, and they transfer geometric structure from the 2D image into the 3D representation. These two functions are entangled within a single attention module. When extending to multi-view conditioning, however, they must be factored apart: a designated \emph{primary view} should anchor the output orientation and establish the corresponding geometric structure, while the remaining \emph{auxiliary views} should only contribute additional geometric structure without altering the established orientation.

We realize this factorization through a dual-stream architecture. Two structurally identical cross-attention modules are maintained: the \emph{primary stream} $\mathrm{CA}_{\mathrm{p}}^{(\ell)}$, initialized from the pretrained weights, which preserves the original dual role described above; and the \emph{auxiliary stream} $\mathrm{CA}_{\mathrm{a}}^{(\ell)}$, initialized by copying all parameters from $\mathrm{CA}_{\mathrm{p}}^{(\ell)}$, which is restricted to geometric structure transfer only. One input view is designated as the primary view $v_p$. The router assigns each 3D latent token to a specific view; tokens assigned to the primary view are processed by $\mathrm{CA}_{\mathrm{p}}^{(\ell)}$, while tokens assigned to any auxiliary view are processed by $\mathrm{CA}_{\mathrm{a}}^{(\ell)}$. Formally, the cross-attention output for token $i$ at layer $\ell$ is:
\begin{equation}
    \mathbf{o}_i^{(\ell)} = 
    \begin{cases}
        \mathrm{CA}_{\mathrm{p}}^{(\ell)}(\tilde{\mathbf{z}}_i^{(\ell)},\; \mathbf{F}_{v_i^{*}}) & \text{if } v_i^{*} = v_p, \\[4pt]
        \mathrm{CA}_{\mathrm{a}}^{(\ell)}(\tilde{\mathbf{z}}_i^{(\ell)},\; \mathbf{F}_{v_i^{*}}) & \text{otherwise},
    \end{cases}
\end{equation}
where $v_i^{*}$ is the view selected by the router (\S\ref{sec:gumbel}) and $\mathbf{F}_{v_i^{*}}$ denotes the patch features of the selected view. To close the gradient path back to the router, the output is modulated by the soft weight from the Gumbel-Softmax:
\begin{equation}
\hat{\mathbf{o}}_i^{(\ell)} = \mathbf{o}_i^{(\ell)} \cdot y_{i,v_i^*},
\end{equation}
where $y_{i,v_i^*} = 1$ in the forward pass (output unchanged) but carries the soft gradient in the backward pass. When only a single view is provided, all tokens are routed to $\mathrm{CA}_{\mathrm{p}}^{(\ell)}$ and the model reduces to standard single-view generation. The only newly introduced parameters are $\mathrm{CA}_{\mathrm{a}}^{(\ell)}$ (a copy of the original module) and the lightweight router (\S\ref{sec:router}).

\subsection{Orientation Perturbation}
\label{sec:perturbation}

While $\mathrm{CA}_{\mathrm{p}}$ closely mirrors the pretrained single-view attention and requires little adaptation, $\mathrm{CA}_{\mathrm{a}}$ needs to decouple orientation control from geometric structure transfer, which requires additional training effort beyond standard fine-tuning.

A key difficulty is that, in standard training, the geometry latent is always canonically aligned with the primary view. Under this setting, $\mathrm{CA}_{\mathrm{a}}$ can take a shortcut by relying on this fixed alignment rather than learning orientation-independent geometric transfer, causing it to fail when the 3D orientation and view directions do not follow the canonical relationship at inference.

To address this, we introduce an orientation perturbation strategy that constructs training samples specifically designed to strengthen $\mathrm{CA}_{\mathrm{a}}$. With probability $p_{\mathrm{pert}}$, the following three operations are jointly applied during training:
\noindent\textbf{(i) Misaligned 3D latent construction.}\; The 3D point cloud is randomly rotated to one of $\{0^{\circ}, 90^{\circ}, 180^{\circ}, 270^{\circ}\}$ in azimuth (with elevation fixed at zero), while explicitly excluding any rotation that would align it with an input view, ensuring the 3D latent is \emph{not} canonically aligned with any conditioning image.
\noindent\textbf{(ii) Primary view removal.}\; The primary view designation is discarded, and all input views are treated as auxiliary, so that every view is processed exclusively through $\mathrm{CA}_{\mathrm{a}}$.
\noindent\textbf{(iii) Selective weight update.}\; Only $\mathrm{CA}_{\mathrm{a}}$ is updated in this mode; $\mathrm{CA}_{\mathrm{p}}$ remains frozen.
\noindent By jointly applying these operations, $\mathrm{CA}_{\mathrm{a}}$ is forced to establish 2D--3D correspondence purely through content-level feature affinity, independent of any orientation prior. % We set $p_{\mathrm{pert}} = 0.2$ in all experiments.

\subsection{Geometric Refinement Stage}
\label{sec:refinement}
To further improve the quality and cleanness of the generated geometry, we adopt a second-stage refinement model following LATTICE~\cite{lai2025lattice}. This refinement stage uses the same multi-view conditioning architecture and training strategy described above---including the token-wise view router, dual-stream attention, and orientation perturbation---applied to a higher-resolution geometric latent space. The coarse geometry produced by the first stage provides positional encoding for the refinement model through a voxel grid, guiding the generation of fine-grained geometric details.

\section{Experiments}
\label{sec:experiments}

\subsection{Experimental Setup}
\label{sec:exp_setup}

\subsubsection{Dataset.}
We train and evaluate \methodName on a filtered subset of Objaverse~\cite{deitke2023objaverse} and ObjaverseXL~\cite{deitke2024objaversexl}, comprising approximately 300K high-quality 3D objects. We follow the data processing pipeline of Hunyuan3D 2.1~\cite{hunyuan3d2025hunyuan3d21imageshighfidelity}, including watertight mesh conversion, point cloud sampling with normals, and conditional image rendering. For rendering, we divide the azimuth range into four bins centered at $\{0^{\circ}, 90^{\circ}, 180^{\circ}, 270^{\circ}\}$, each spanning $\pm 45^{\circ}$. Within each bin, we render 16 images with randomly sampled field of view ($20^{\circ}$--$70^{\circ}$) and elevation ($-90^{\circ}$--$90^{\circ}$). We hold out 100 objects for evaluation that are not seen during training. 

To assess generalization to real-world captures, we construct a multi-source evaluation set of 200 objects, \textsc{Anyview-200}. This set comprises multi-view images generated by Nano Banana Pro~\cite{team2023gemini} (which may contain cross-view inconsistencies inherent to AI generation), photographs captured from real-world scenes, and images collected from the internet. The number of input views per object ranges from 1 to 12, and the views exhibit diverse real-world challenges including strong perspective distortion, varying camera intrinsics across views of the same object, and incomplete coverage of the object surface.

\subsubsection{Evaluation Metrics.}
We adopt a comprehensive set of metrics spanning both geometric accuracy and semantic quality. For geometry, we report Chamfer Distance (CD, $\times 10^{-3}$) and F-Score (F1, \%) at a threshold of 0.1 and 0.05. To assess the semantic quality of generated geometry, we report ULIP~\cite{Xue_2023_CVPR} and Uni3D~\cite{zhou2023uni3d} image similarities (ULIP-I and Uni-I), which measure the alignment between the generated 3D shape and the input image condition in a shared embedding space.

\subsubsection{Baseline Methods.}
We compare \methodName against the following categories of methods: (1) \emph{Single-view generative models}: Hunyuan3D 2.1~\cite{hunyuan3d2025hunyuan3d21imageshighfidelity}, Trellis~\cite{xiang2024trellis}, Trellis2~\cite{xiang2025trellis2}; (2) \emph{Multi-view reconstruction methods}: LGM~\cite{tang2025lgm} and VGGT~\cite{wang2025vggt}; (3) \emph{Multi-view conditioned generation}: ReconViaGen~\cite{chang2025reconviagenaccuratemultiview3d}. We additionally include qualitative comparisons with two leading commercial 3D generation services (denoted as Model~1 and Model~2 in \cref{fig:qualitative}) to demonstrate the competitiveness of \methodName against state-of-the-art industrial systems. For single-view baselines, we provide the same reference view used by \methodName. For multi-view methods, we provide all $V$ available views.

\subsubsection{Implementation Details.}
\paragraph{First Stage (Coarse Geometry).}
We initialize the backbone 3DShape2VecSet VAE and DiT from the pretrained weights of Hunyuan3D 2.1~\cite{hunyuan3d2025hunyuan3d21imageshighfidelity}. The DiT has a hidden dimension of 2048 and operates on 4096 latent tokens. The image features are extracted from a frozen DINOv2 ViT-L/14~\cite{oquab2024dinov2} encoder, yielding $S{=}256$ patch tokens of dimension $D{=}1024$ per view, which are projected to match the DiT hidden dimension. The router projection matrices $\mathbf{W}_q^{(\ell)}, \mathbf{W}_k^{(\ell)}$ and the auxiliary attention path $\mathrm{CA}_{\mathrm{a}}$ are initialized by copying the corresponding pretrained weights. The Gumbel-Softmax temperature $\tau$ is set to 1.0 throughout training. During training, we randomly sample 1 to 4 auxiliary views per instance with uniform probability, allowing the model to handle variable view counts. We train with the AdamW optimizer~\cite{loshchilov2017decoupled} at a learning rate of $1 \times 10^{-5}$ with a cosine decay schedule. The orientation perturbation probability is set to $p_{\mathrm{pert}} = 0.2$. The model is trained for 50K iterations with a per-GPU batch size of 4, requiring approximately 3$\sim$4 GPU days. At test time, the model generalizes to an arbitrary number of views without modification, supporting scaling well beyond the training range (see supplementary for more comparison).

% \paragraph{Second Stage (Geometric Refinement).}
% We adopt the LATTICE~\cite{lai2025lattice} architecture for higher-resolution geometry refinement, restructured to follow the same DiT-based design as the first stage but with increased depth and a hidden dimension of 2048, resulting in a 1.6B-parameter model operating on 6144 latent tokens. The image encoder is upgraded to a frozen DINOv2 ViT-G/14~\cite{oquab2024dinov2}, producing patch tokens of dimension $D{=}1536$. Positional encoding is applied to the 3D voxel queries to encode their spatial positions. To enable seamless integration of the token-wise view router, we replace the original self-attention-based conditioning in LATTICE with cross-attention. Our multi-view conditioning modules---the view router, dual-stream attention, and orientation perturbation strategy---are then integrated into this stage following the same design as the first stage, with the same 1-to-4 random view sampling strategy during training. We strictly follow the training recipe of LATTICE, including its data pipeline, loss formulation, and scheduling, with the model initialized from the pretrained LATTICE weights. This stage requires approximately 10 GPU days.
\paragraph{Second Stage (Geometric Refinement).}
We build upon LATTICE~\cite{lai2025lattice} with a 1.6B-parameter DiT operating on 6144 latent tokens, replacing its self-attention-based conditioning with cross-attention to enable integration of our multi-view modules (view router, dual-stream attention, and orientation perturbation). The image encoder is upgraded to DINOv2 ViT-G/14~\cite{oquab2024dinov2}; all other training settings follow LATTICE. This stage requires approximately 10 GPU days.

\subsection{Comparison}
\label{sec:comparison}
\subsubsection{Quantitative Results.}
\cref{tab:main_comparison} summarizes results on our held-out test set. Single-view methods (Hunyuan3D 2.0, Trellis2) achieve reasonable quality but suffer from high CD due to hallucinated geometry in unseen regions. Multi-view reconstruction methods reduce this ambiguity yet remain limited: LGM produces over-smoothed outputs with the lowest F1 scores across both thresholds, and VGGT, despite better geometric coverage, scores poorly on semantic metrics (ULIP-I: 0.103) due to the absence of a generative prior. Among multi-view conditioned methods, Hunyuan3D 2.0-MV shows marginal improvement over its single-view counterpart, and ReconViaGen achieves the best baseline performance with the lowest CD (32.055) and highest semantic scores (Uni-I: 0.345), though the geometric gap over single-view methods remains modest. ROAR-3D stage~1 already outperforms all baselines across every metric (CD: 25.372 vs.\ 32.055 for ReconViaGen), and stage~2 refinement brings further substantial improvement, reducing CD to 21.039 and raising F1(0.05) to 81.6---a 34\% CD reduction and 12\% F1(0.05) improvement over the strongest baseline.

\begin{table}[tbp]
  \caption{Quantitative comparison on our held-out test set. Best results are in \textbf{bold}, second best are \underline{underlined}. CD is multiplied by $10^{3}$ (lower is better). Higher is better for all other metrics.}
  \label{tab:main_comparison}
  \centering
  \resizebox{0.85\linewidth}{!}{
  \setlength{\tabcolsep}{5pt}
  \begin{tabular}{l c c c c c}
    \toprule
    Method & CD$\downarrow$ & F1(0.1)$\uparrow$ & F1(0.05)$\uparrow$ & ULIP-I$\uparrow$ & Uni-I$\uparrow$ \\
    \midrule
    \multicolumn{6}{l}{\emph{Single-view generative models}} \\
    Hunyuan3D 2.0~\cite{hunyuan3d2025hunyuan3d21imageshighfidelity}       & 35.421 & 80.5 & 64.2 & 0.141 & 0.302 \\
    Hunyuan3D 2.1~\cite{hunyuan3d2025hunyuan3d21imageshighfidelity}       & 34.031 & 82.6 & 68.7 & 0.147 & 0.317 \\
    Trellis2~\cite{xiang2025trellis2}       & 33.722 & 84.4 & 71.0 & 0.149 & 0.321 \\
    \midrule
    \multicolumn{6}{l}{\emph{Multi-view reconstruction}} \\
    LGM~\cite{tang2025lgm}                  & 58.009 & 71.0 & 52.3 & 0.089 & 0.215 \\
    VGGT~\cite{wang2025vggt}                & 34.686 & 81.3 & 65.4 & 0.103 & 0.211 \\
    \midrule
    \multicolumn{6}{l}{\emph{Multi-view conditioned generation}} \\
    Hunyuan3D 2.0-MV~\cite{hunyuan3d2025hunyuan3d21imageshighfidelity}     & 33.954 & 82.4 & 67.1 & 0.147 & 0.301 \\
    ReconViaGen~\cite{chang2025reconviagenaccuratemultiview3d} & 32.055 & 85.1 & 72.7 & 0.165 & 0.345 \\
    \textbf{ROAR-3D (Ours) - stage 1}        & \underline{25.372} & \underline{88.7} & \underline{76.8} & \underline{0.168} & \underline{0.346} \\
    \textbf{ROAR-3D (Ours) - stage 2}        & \textbf{21.039} & \textbf{91.2} & \textbf{81.6} & \textbf{0.173} & \textbf{0.349} \\
    \bottomrule
  \end{tabular}
  }
\end{table}

\subsubsection{Qualitative Results.}
\begin{figure}[tbp]
  \centering
  \includegraphics[width=\linewidth]{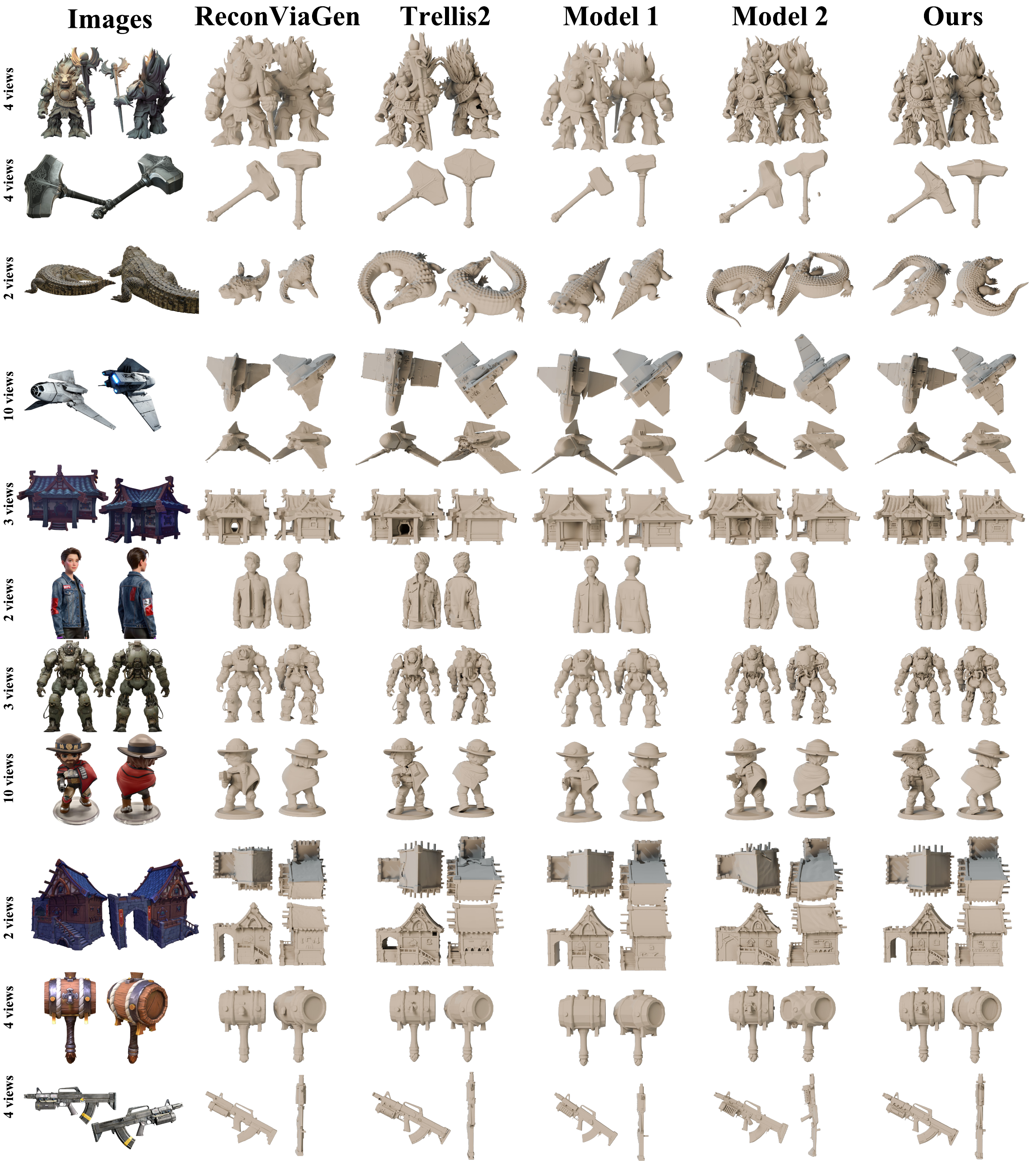}
  \caption{Qualitative comparison with baseline methods. \methodName produces complete, high-fidelity 3D shapes that are consistent with all input views.}
  \label{fig:qualitative}
\end{figure}

\cref{fig:qualitative} presents qualitative comparisons across representative examples. Several trends are visible. First, \methodName produces geometrically correct structures where other methods falter: the hammer (row~2) maintains its correct upright orientation, the aircraft (row~4) faithfully reproduces both the nose and tail geometry, and later rows show that our outputs consistently avoid the distorted or tilted shapes that appear in some baselines. Second, our method achieves higher detail fidelity with the input views: the warrior (row~1) preserves fine-grained surface details that match the input, the portrait (row~6) correctly reconstructs the pendant on the back of the jacket visible in the reference image, and the cartoon characters (rows~7--8) retain ornamental details consistent across views. Third, \methodName demonstrates robustness to inconsistent inputs: the crocodile (row~3) is captured from views with noticeable appearance discrepancies, yet our method produces a physically plausible and complete geometry while maximizing fidelity to each input view. Overall, by combining the generative prior of the pretrained DiT with accurate multi-view conditioning, \methodName yields complete 3D shapes with correct global orientation and faithful local details, outperforming both reconstruction-based and generation-based baselines.

\subsection{Ablation Study}
\label{sec:ablation}

\begin{figure}[tbp]
  \centering
  \includegraphics[width=\linewidth]{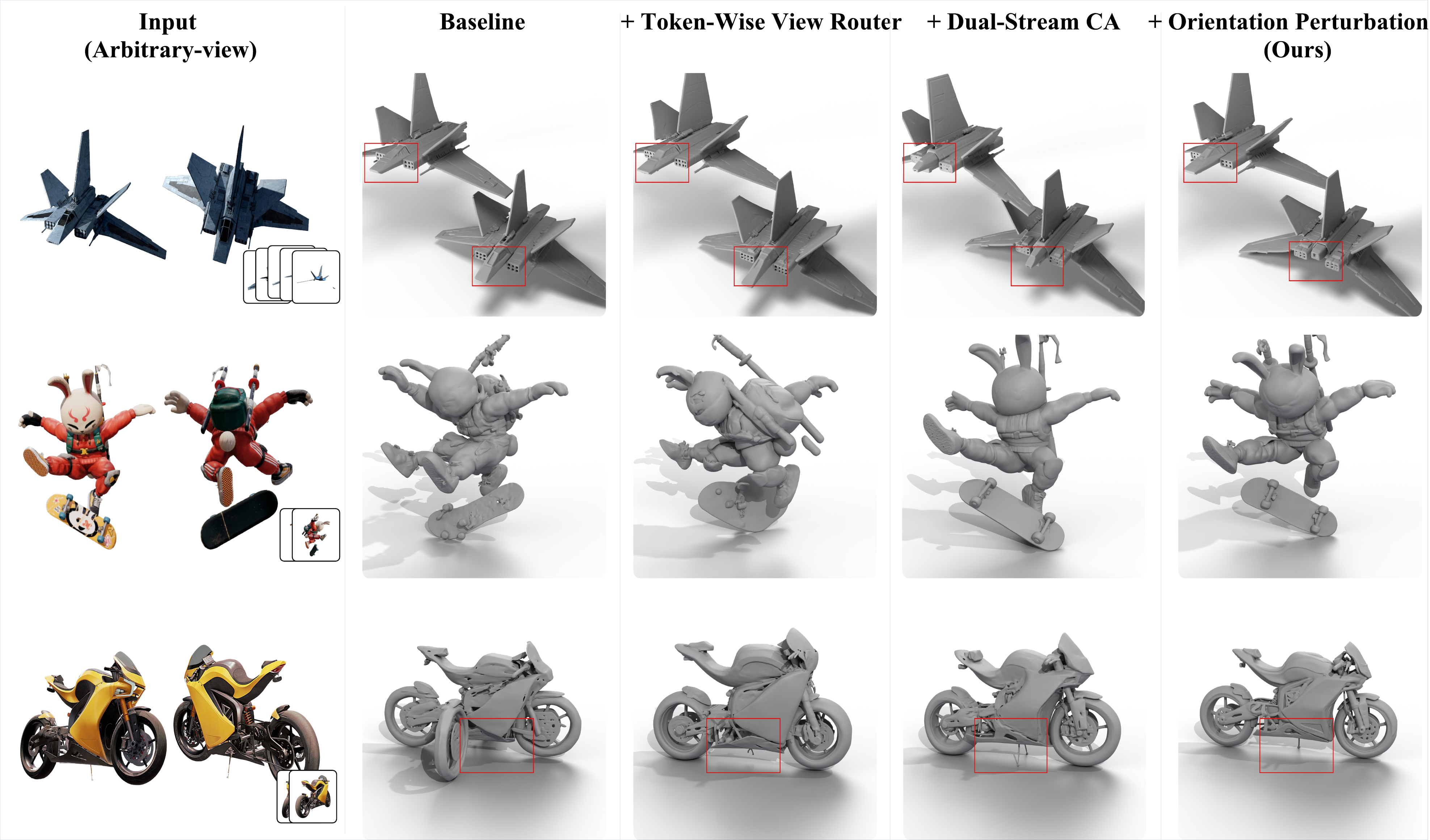}
  \caption{Ablation study on the three key components of \methodName. Removing any single component leads to incorrect orientation or degraded geometry (highlighted with red boxes). Best viewed when zoomed in.}
  \label{fig:ablation_qualitative}
\end{figure}

We conduct ablation studies on the \textsc{Anyview-200} dataset to validate the key design choices of \methodName. To isolate the effect of our multi-view conditioning module, all ablations are performed using the first-stage model only (without the lattice refinement stage). Unless otherwise stated, all variants are trained under identical settings for 50K iterations.

\subsubsection{Component Ablation.}
We incrementally add each proposed component to a baseline that simply concatenates all multi-view tokens and feeds them into the original attention layers. The three components are added in order: (1) the \emph{token-wise view router}, which replaces naive concatenation with per-token view selection; (2) \emph{dual-stream attention}, which separates the primary-view path from the auxiliary-view path; and (3) the \emph{orientation perturbation strategy}, which yields our full model. Quantitative results are reported in \cref{tab:ablation}, and qualitative comparisons are shown in \cref{fig:ablation_qualitative}.

As shown in \cref{fig:ablation_qualitative}, each added component brings visible improvement in orientation accuracy and geometric fidelity. For the fighter jet (row~1), the baseline incorrectly replicates the nose geometry onto the tail; introducing the router partially alleviates this, and adding dual-stream attention further improves structural separation, but only the full model with orientation perturbation correctly distinguishes front and rear. For the cartoon character (row~2), the baseline produces both incorrect orientation and degraded details; each component progressively improves orientation control and geometric faithfulness, with the full model being the only variant that achieves both. For the motorcycle (row~3), the baseline generates duplicated wheels; the router resolves this, but the kickstand geometry remains incorrect until dual-stream attention and orientation perturbation are both applied. We note that some intermediate variants still exhibit orientation errors; for fair visual comparison, we manually rotate them to approximately matching viewpoints.

\begin{table}[tbp]
  \caption{Ablation on the three key components of \methodName.}
  \label{tab:ablation}
  \centering
  \setlength{\tabcolsep}{4pt}
  \begin{tabular}{l c c c c c}
    \toprule
    Variant & CD$\downarrow$ & F1(0.1)$\uparrow$ & F1(0.05)$\uparrow$ & ULIP-I$\uparrow$ & Uni-I$\uparrow$ \\
    \midrule
    Baseline                                       & 39.412 & 79.6 & 66.3 & 0.154 & 0.322 \\
    $+$ Token-wise View Router                     & 28.143 & 86.3 & 73.5 & 0.159 & 0.335 \\
    $+$ Dual-Stream Attention                      & 26.732 & 87.5 & 75.1 & 0.163 & 0.339 \\
    \midrule
    $+$ Orientation Perturbation (Ours)            & \textbf{25.372} & \textbf{88.7} & \textbf{76.8} & \textbf{0.168} & \textbf{0.346} \\
    \bottomrule
  \end{tabular}
\end{table}

\section{Conclusion}
\label{sec:conclusion}
We presented ROAR-3D, a lightweight method that upgrades pretrained single-view 3D generative models to accept arbitrary unposed multi-view input. Three complementary components---a token-wise view router, dual-stream attention, and orientation perturbation---decouple orientation control from geometry transfer, enabling effective multi-view fusion without external modules, camera poses, or additional inference cost. ROAR-3D achieves state-of-the-art quality and supports test-time view scaling from 1 to 12+ views.

% \section*{Acknowledgements}
% Please insert your acknowledgments here.

\bibliographystyle{splncs04}
\bibliography{main}

@String(TOG   = {ACM Transactions on Graphics (TOG)})

@String(CVPR  = {Proceedings of the IEEE/CVF Conference on Computer Vision and Pattern Recognition (CVPR)})

@String(ICCV  = {Proceedings of the IEEE/CVF International Conference on Computer Vision (ICCV)})

@String(ECCV  = {Proceedings of the European Conference on Computer Vision (ECCV)})

@String(NeurIPS  = {Proceedings of the Advances in Neural Information Processing Systems (NeurIPS)})

@String(ICLR  = {Proceedings of the The International Conference on Learning Representations (ICLR)})

@string(SIGGRAPH  = {Proceedings of the Annual Conference on Computer Graphics and Interactive Techniques (SIGGRAPH)})

@string(SIGGRAPHA = {Proceedings of the ACM SIGGRAPH Conference and Exhibition on Computer Graphics and Interactive Techniques in Asia (SIGGRAPH Aisa)})

@String(TOG   = {TOG})

@String(CVPR  = {CVPR})

@String(ICCV  = {ICCV})

@String(ECCV  = {ECCV})

@String(NeurIPS  = {NeurIPS})

@String(ICLR  = {ICLR})

@string(SIGGRAPH  = {SIGGRAPH})

@string(SIGGRAPHA = {SIGGRAPH Aisa})

@inproceedings{mildenhall2020nerf,
  title={NeRF: Representing Scenes as Neural Radiance Fields for View Synthesis},
  author={Mildenhall, Ben and Srinivasan, Pratul P and Tancik, Matthew and Barron, Jonathan T and Ramamoorthi, Ravi and Ng, Ren},
  booktitle=ECCV,
  pages={405--421},
  year={2020}
}

@article{kerbl20233dgs,
  title={3d gaussian splatting for real-time radiance field rendering.},
  author={Kerbl, Bernhard and Kopanas, Georgios and Leimk{\"u}hler, Thomas and Drettakis, George},
  journal=TOG,
  volume={42},
  number={4},
  pages={139--1},
  year={2023}
}

@article{wang2025vggt,
  title={Vggt: Visual geometry grounded transformer},
  author={Wang, Jianyuan and Chen, Minghao and Karaev, Nikita and Vedaldi, Andrea and Rupprecht, Christian and Novotny, David},
  journal={arXiv preprint arXiv:2503.11651},
  year={2025}
}

@inproceedings{luo2021diffusion,
  title={Diffusion probabilistic models for 3d point cloud generation},
  author={Luo, Shitong and Hu, Wei},
  booktitle=CVPR,
  pages={2837--2845},
  year={2021}
}

@inproceedings{zhou20213d,
  title={3d shape generation and completion through point-voxel diffusion},
  author={Zhou, Linqi and Du, Yilun and Wu, Jiajun},
  booktitle=ICCV,
  pages={5826--5835},
  year={2021}
}

@inproceedings{hui2022neural,
  title={Neural wavelet-domain diffusion for 3d shape generation},
  author={Hui, Ka-Hei and Li, Ruihui and Hu, Jingyu and Fu, Chi-Wing},
  booktitle=SIGGRAPHA,
  pages={1--9},
  year={2022}
}

@inproceedings{shue2023triplanediffusion,
  title={3d neural field generation using triplane diffusion},
  author={Shue, J Ryan and Chan, Eric Ryan and Po, Ryan and Ankner, Zachary and Wu, Jiajun and Wetzstein, Gordon},
  booktitle=CVPR,
  pages={20875--20886},
  year={2023}
}

@article{zhang2024gaussiancube,
  title={Gaussiancube: Structuring gaussian splatting using optimal transport for 3d generative modeling},
  author={Zhang, Bowen and Cheng, Yiji and Yang, Jiaolong and Wang, Chunyu and Zhao, Feng and Tang, Yansong and Chen, Dong and Guo, Baining},
  journal={arXiv e-prints},
  pages={arXiv--2403},
  year={2024}
}

@article{ye2025hi3dgen,
  title={Hi3dgen: High-fidelity 3d geometry generation from images via normal bridging},
  author={Ye, Chongjie and Wu, Yushuang and Lu, Ziteng and Chang, Jiahao and Guo, Xiaoyang and Zhou, Jiaqing and Zhao, Hao and Han, Xiaoguang},
  journal={arXiv preprint arXiv:2503.22236},
  volume={3},
  year={2025}
}

@article{zhang2024clay,
  title={CLAY: A Controllable Large-scale Generative Model for Creating High-quality 3D Assets},
  author={Zhang, Longwen and Wang, Ziyu and Zhang, Qixuan and Qiu, Qiwei and Pang, Anqi and Jiang, Haoran and Yang, Wei and Xu, Lan and Yu, Jingyi},
  journal=TOG,
  volume={43},
  number={4},
  pages={1--20},
  year={2024},
  publisher={ACM New York, NY, USA}
}

@article{wu2024direct3d,
  title={Direct3D: Scalable Image-to-3D Generation via 3D Latent Diffusion Transformer},
  author={Wu, Shuang and Lin, Youtian and Zhang, Feihu and Zeng, Yifei and Xu, Jingxi and Torr, Philip and Cao, Xun and Yao, Yao},
  journal={arXiv preprint arXiv:2405.14832},
  year={2024}
}

@article{li2024craftsman,
  title={CraftsMan: High-fidelity Mesh Generation with 3D Native Generation and Interactive Geometry Refiner},
  author={Li, Weiyu and Liu, Jiarui and Chen, Rui and Liang, Yixun and Chen, Xuelin and Tan, Ping and Long, Xiaoxiao},
  journal={arXiv preprint arXiv:2405.14979},
  year={2024}
}

@inproceedings{honglrm,
  title={LRM: Large Reconstruction Model for Single Image to 3D},
  author={Hong, Yicong and Zhang, Kai and Gu, Jiuxiang and Bi, Sai and Zhou, Yang and Liu, Difan and Liu, Feng and Sunkavalli, Kalyan and Bui, Trung and Tan, Hao},
  booktitle=ICLR,
  year={2023}
}

@article{poole2022dreamfusion,
  title={Dreamfusion: Text-to-3d using 2d diffusion},
  author={Poole, Ben and Jain, Ajay and Barron, Jonathan T and Mildenhall, Ben},
  journal={arXiv preprint arXiv:2209.14988},
  year={2022}
}

@article{liu2024one2345,
  title={One-2-3-45: Any single image to 3d mesh in 45 seconds without per-shape optimization},
  author={Liu, Minghua and Xu, Chao and Jin, Haian and Chen, Linghao and Varma T, Mukund and Xu, Zexiang and Su, Hao},
  journal=NeurIPS,
  volume={36},
  year={2023}
}

@inproceedings{liu2024one2345pp,
  title={One-2-3-45++: Fast single image to 3d objects with consistent multi-view generation and 3d diffusion},
  author={Liu, Minghua and Shi, Ruoxi and Chen, Linghao and Zhang, Zhuoyang and Xu, Chao and Wei, Xinyue and Chen, Hansheng and Zeng, Chong and Gu, Jiayuan and Su, Hao},
  booktitle=CVPR,
  pages={10072--10083},
  year={2024}
}

@inproceedings{dreamgaussian,
  title={DreamGaussian: Generative Gaussian Splatting for Efficient 3D Content Creation},
  author={Tang, Jiaxiang and Ren, Jiawei and Zhou, Hang and Liu, Ziwei and Zeng, Gang},
  booktitle=ICLR,
  year={2024}
}

@article{wu2024unique3d,
  title={Unique3D: High-Quality and Efficient 3D Mesh Generation from a Single Image},
  author={Wu, Kailu and Liu, Fangfu and Cai, Zhihan and Yan, Runjie and Wang, Hanyang and Hu, Yating and Duan, Yueqi and Ma, Kaisheng},
  journal={arXiv preprint arXiv:2405.20343},
  year={2024}
}

@inproceedings{qiu2024richdreamer,
  title={Richdreamer: A generalizable normal-depth diffusion model for detail richness in text-to-3d},
  author={Qiu, Lingteng and Chen, Guanying and Gu, Xiaodong and Zuo, Qi and Xu, Mutian and Wu, Yushuang and Yuan, Weihao and Dong, Zilong and Bo, Liefeng and Han, Xiaoguang},
  booktitle=CVPR,
  pages={9914--9925},
  year={2024}
}

@article{wang2024prolificdreamer,
  title={Prolificdreamer: High-fidelity and diverse text-to-3d generation with variational score distillation},
  author={Wang, Zhengyi and Lu, Cheng and Wang, Yikai and Bao, Fan and Li, Chongxuan and Su, Hang and Zhu, Jun},
  journal=NeurIPS,
  volume={36},
  year={2024}
}

@inproceedings{lin2023magic3d,
  title={Magic3d: High-resolution text-to-3d content creation},
  author={Lin, Chen-Hsuan and Gao, Jun and Tang, Luming and Takikawa, Towaki and Zeng, Xiaohui and Huang, Xun and Kreis, Karsten and Fidler, Sanja and Liu, Ming-Yu and Lin, Tsung-Yi},
  booktitle=CVPR,
  pages={300--309},
  year={2023}
}

@article{nichol2022pointe,
  title={Point-e: A system for generating 3d point clouds from complex prompts},
  author={Nichol, Alex and Jun, Heewoo and Dhariwal, Prafulla and Mishkin, Pamela and Chen, Mark},
  journal={arXiv preprint arXiv:2212.08751},
  year={2022}
}

@inproceedings{tang2025lgm,
  title={Lgm: Large multi-view gaussian model for high-resolution 3d content creation},
  author={Tang, Jiaxiang and Chen, Zhaoxi and Chen, Xiaokang and Wang, Tengfei and Zeng, Gang and Liu, Ziwei},
  booktitle=ECCV,
  pages={1--18},
  year={2025},
  organization={Springer}
}

@article{xu2024grm,
  title={Grm: Large gaussian reconstruction model for efficient 3d reconstruction and generation},
  author={Xu, Yinghao and Shi, Zifan and Yifan, Wang and Chen, Hansheng and Yang, Ceyuan and Peng, Sida and Shen, Yujun and Wetzstein, Gordon},
  journal={arXiv preprint arXiv:2403.14621},
  year={2024}
}

@inproceedings{wang2025crm,
  title={Crm: Single image to 3d textured mesh with convolutional reconstruction model},
  author={Wang, Zhengyi and Wang, Yikai and Chen, Yifei and Xiang, Chendong and Chen, Shuo and Yu, Dajiang and Li, Chongxuan and Su, Hang and Zhu, Jun},
  booktitle=ECCV,
  pages={57--74},
  year={2025},
  organization={Springer}
}

@article{xu2024instantmesh,
  title={Instantmesh: Efficient 3d mesh generation from a single image with sparse-view large reconstruction models},
  author={Xu, Jiale and Cheng, Weihao and Gao, Yiming and Wang, Xintao and Gao, Shenghua and Shan, Ying},
  journal={arXiv preprint arXiv:2404.07191},
  year={2024}
}

@article{li2023instant3d,
  title={Instant3d: Fast text-to-3d with sparse-view generation and large reconstruction model},
  author={Li, Jiahao and Tan, Hao and Zhang, Kai and Xu, Zexiang and Luan, Fujun and Xu, Yinghao and Hong, Yicong and Sunkavalli, Kalyan and Shakhnarovich, Greg and Bi, Sai},
  journal={arXiv preprint arXiv:2311.06214},
  year={2023}
}

@article{wei2024meshlrm,
  title={Meshlrm: Large reconstruction model for high-quality mesh},
  author={Wei, Xinyue and Zhang, Kai and Bi, Sai and Tan, Hao and Luan, Fujun and Deschaintre, Valentin and Sunkavalli, Kalyan and Su, Hao and Xu, Zexiang},
  journal={arXiv preprint arXiv:2404.12385},
  year={2024}
}

@article{zhang20233dshape2vecset,
  title={3dshape2vecset: A 3d shape representation for neural fields and generative diffusion models},
  author={Zhang, Biao and Tang, Jiapeng and Niessner, Matthias and Wonka, Peter},
  journal=TOG,
  volume={42},
  number={4},
  pages={1--16},
  year={2023},
  publisher={ACM New York, NY, USA}
}

@inproceedings{tang2023make,
  title={Make-it-3d: High-fidelity 3d creation from a single image with diffusion prior},
  author={Tang, Junshu and Wang, Tengfei and Zhang, Bo and Zhang, Ting and Yi, Ran and Ma, Lizhuang and Chen, Dong},
  booktitle=ICCV,
  pages={22819--22829},
  year={2023}
}

@article{li2024era3d,
  title={Era3D: High-Resolution Multiview Diffusion using Efficient Row-wise Attention},
  author={Li, Peng and Liu, Yuan and Long, Xiaoxiao and Zhang, Feihu and Lin, Cheng and Li, Mengfei and Qi, Xingqun and Zhang, Shanghang and Luo, Wenhan and Tan, Ping and others},
  journal={arXiv preprint arXiv:2405.11616},
  year={2024}
}

@inproceedings{muller2023diffrf,
  title={Diffrf: Rendering-guided 3d radiance field diffusion},
  author={M{\"u}ller, Norman and Siddiqui, Yawar and Porzi, Lorenzo and Bulo, Samuel Rota and Kontschieder, Peter and Nie{\ss}ner, Matthias},
  booktitle=CVPR,
  pages={4328--4338},
  year={2023}
}

@inproceedings{chen2023single,
  title={Single-stage diffusion nerf: A unified approach to 3d generation and reconstruction},
  author={Chen, Hansheng and Gu, Jiatao and Chen, Anpei and Tian, Wei and Tu, Zhuowen and Liu, Lingjie and Su, Hao},
  booktitle=ICCV,
  pages={2416--2425},
  year={2023}
}

@inproceedings{wang2023rodin,
  title={Rodin: A generative model for sculpting 3d digital avatars using diffusion},
  author={Wang, Tengfei and Zhang, Bo and Zhang, Ting and Gu, Shuyang and Bao, Jianmin and Baltrusaitis, Tadas and Shen, Jingjing and Chen, Dong and Wen, Fang and Chen, Qifeng and others},
  booktitle=CVPR,
  pages={4563--4573},
  year={2023}
}

@article{zhao2024michelangelo,
  title={Michelangelo: Conditional 3d shape generation based on shape-image-text aligned latent representation},
  author={Zhao, Zibo and Liu, Wen and Chen, Xin and Zeng, Xianfang and Wang, Rui and Cheng, Pei and Fu, Bin and Chen, Tao and Yu, Gang and Gao, Shenghua},
  journal=NeurIPS,
  volume={36},
  year={2024}
}

@article{tang2023volumediffusion,
  title={Volumediffusion: Flexible text-to-3d generation with efficient volumetric encoder},
  author={Tang, Zhicong and Gu, Shuyang and Wang, Chunyu and Zhang, Ting and Bao, Jianmin and Chen, Dong and Guo, Baining},
  journal={arXiv preprint arXiv:2312.11459},
  year={2023}
}

@inproceedings{xudmv3d,
  title={DMV3D: Denoising Multi-view Diffusion Using 3D Large Reconstruction Model},
  author={Xu, Yinghao and Tan, Hao and Luan, Fujun and Bi, Sai and Wang, Peng and Li, Jiahao and Shi, Zifan and Sunkavalli, Kalyan and Wetzstein, Gordon and Xu, Zexiang and others},
  booktitle=ICLR,
  year={2024}
}

@article{zuo2024videomv,
  title={Videomv: Consistent multi-view generation based on large video generative model},
  author={Zuo, Qi and Gu, Xiaodong and Qiu, Lingteng and Dong, Yuan and Zhao, Zhengyi and Yuan, Weihao and Peng, Rui and Zhu, Siyu and Dong, Zilong and Bo, Liefeng and others},
  journal={arXiv preprint arXiv:2403.12010},
  year={2024}
}

@inproceedings{deitke2023objaverse,
  title={Objaverse: A universe of annotated 3d objects},
  author={Deitke, Matt and Schwenk, Dustin and Salvador, Jordi and Weihs, Luca and Michel, Oscar and VanderBilt, Eli and Schmidt, Ludwig and Ehsani, Kiana and Kembhavi, Aniruddha and Farhadi, Ali},
  booktitle=ICCV,
  pages={13142--13153},
  year={2023}
}

@article{deitke2024objaversexl,
  title={Objaverse-xl: A universe of 10m+ 3d objects},
  author={Deitke, Matt and Liu, Ruoshi and Wallingford, Matthew and Ngo, Huong and Michel, Oscar and Kusupati, Aditya and Fan, Alan and Laforte, Christian and Voleti, Vikram and Gadre, Samir Yitzhak and others},
  journal=NeurIPS,
  volume={36},
  year={2024}
}

@article{xiang2024trellis,
  title={Structured 3d latents for scalable and versatile 3d generation},
  author={Xiang, Jianfeng and Lv, Zelong and Xu, Sicheng and Deng, Yu and Wang, Ruicheng and Zhang, Bowen and Chen, Dong and Tong, Xin and Yang, Jiaolong},
  journal={arXiv preprint arXiv:2412.01506},
  year={2024}
}

@article{li2025triposg,
  title={TripoSG: High-Fidelity 3D Shape Synthesis using Large-Scale Rectified Flow Models},
  author={Li, Yangguang and Zou, Zi-Xin and Liu, Zexiang and Wang, Dehu and Liang, Yuan and Yu, Zhipeng and Liu, Xingchao and Guo, Yuan-Chen and Liang, Ding and Ouyang, Wanli and others},
  journal={arXiv preprint arXiv:2502.06608},
  year={2025}
}

@inproceedings{wang2024dust3r,
  title={Dust3r: Geometric 3d vision made easy},
  author={Wang, Shuzhe and Leroy, Vincent and Cabon, Yohann and Chidlovskii, Boris and Revaud, Jerome},
  booktitle=CVPR,
  pages={20697--20709},
  year={2024}
}

@article{oquab2024dinov2,
  title={DINOv2: Learning Robust Visual Features without Supervision},
  author={Oquab, Maxime and Darcet, Timoth{\'e}e and Moutakanni, Th{\'e}o and Vo, Huy and Szafraniec, Marc and Khalidov, Vasil and Fernandez, Pierre and Haziza, Daniel and Massa, Francisco and El-Nouby, Alaaeldin and others},
  journal={Transactions on Machine Learning Research Journal},
  pages={1--31},
  year={2024}
}

@article{zhao2025hunyuan3d2,
  title={Hunyuan3D 2.0: Scaling Diffusion Models for High Resolution Textured 3D Assets Generation},
  author={Zhao, Zibo and Lai, Zeqiang and Lin, Qingxiang and Zhao, Yunfei and Liu, Haolin and Yang, Shuhui and Feng, Yifei and Yang, Mingxin and Zhang, Sheng and Yang, Xianghui and others},
  journal={arXiv preprint arXiv:2501.12202},
  year={2025}
}

@inproceedings{chen2023text2tex,
  title={Text2tex: Text-driven texture synthesis via diffusion models},
  author={Chen, Dave Zhenyu and Siddiqui, Yawar and Lee, Hsin-Ying and Tulyakov, Sergey and Nie{\ss}ner, Matthias},
  booktitle={Proceedings of the IEEE/CVF international conference on computer vision},
  pages={18558--18568},
  year={2023}
}

@inproceedings{liu2024text,
  title={Text-guided texturing by synchronized multi-view diffusion},
  author={Liu, Yuxin and Xie, Minshan and Liu, Hanyuan and Wong, Tien-Tsin},
  booktitle={SIGGRAPH Asia 2024 Conference Papers},
  pages={1--11},
  year={2024}
}

@inproceedings{feng2025romantex,
  title={Romantex: Decoupling 3d-aware rotary positional embedded multi-attention network for texture synthesis},
  author={Feng, Yifei and Yang, Mingxin and Yang, Shuhui and Zhang, Sheng and Yu, Jiaao and Zhao, Zibo and Liu, Yuhong and Jiang, Jie and Guo, Chunchao},
  booktitle={Proceedings of the IEEE/CVF International Conference on Computer Vision},
  pages={17203--17213},
  year={2025}
}

@inproceedings{he2025materialmvp,
  title={Materialmvp: Illumination-invariant material generation via multi-view pbr diffusion},
  author={He, Zebin and Yang, Mingxin and Yang, Shuhui and Tang, Yixuan and Wang, Tao and Zhang, Kaihao and Chen, Guanying and Liu, Yuhong and Jiang, Jie and Guo, Chunchao and others},
  booktitle={Proceedings of the IEEE/CVF International Conference on Computer Vision},
  pages={26294--26305},
  year={2025}
}

@article{luo2025matpedia,
  title={MatPedia: A Universal Generative Foundation for High-Fidelity Material Synthesis},
  author={Luo, Di and Yang, Shuhui and Yang, Mingxin and Lu, Jiawei and Tang, Yixuan and Han, Xintong and Chen, Zhuo and Wang, Beibei and Guo, Chunchao},
  journal={arXiv preprint arXiv:2511.16957},
  year={2025}
}

@article{loshchilov2017decoupled,
  title={Decoupled weight decay regularization},
  author={Loshchilov, Ilya and Hutter, Frank},
  journal={arXiv preprint arXiv:1711.05101},
  year={2017}
}

@misc{hunyuan3d,
  author = {{Tencent Hunyuan}},
  title = {Hunyuan3D},
  howpublished = {https://3d.hunyuan.tencent.com/},
  year = {2024},
  urldate = {2026-03-04}
}

@article{furukawa2015multi,
  title={Multi-view stereo: A tutorial},
  author={Furukawa, Yasutaka and Hern{\'a}ndez, Carlos and others},
  journal={Foundations and trends{\textregistered} in Computer Graphics and Vision},
  volume={9},
  number={1-2},
  pages={1--148},
  year={2015},
  publisher={Now Publishers, Inc.}
}

@inproceedings{galliani2015massively,
  title={Massively parallel multiview stereopsis by surface normal diffusion},
  author={Galliani, Silvano and Lasinger, Katrin and Schindler, Konrad},
  booktitle=ICCV,
  pages={873--881},
  year={2015}
}

@inproceedings{schonberger2016pixelwise,
  title={Pixelwise view selection for unstructured multi-view stereo},
  author={Sch{\"o}nberger, Johannes L and Zheng, Enliang and Frahm, Jan-Michael and Pollefeys, Marc},
  booktitle=ECCV,
  pages={501--518},
  year={2016},
  organization={Springer}
}

@inproceedings{xu2019multi,
  title={Multi-scale geometric consistency guided multi-view stereo},
  author={Xu, Qingshan and Tao, Wenbing},
  booktitle=CVPR,
  pages={5483--5492},
  year={2019}
}

@inproceedings{yao2018mvsnet,
  title={Mvsnet: Depth inference for unstructured multi-view stereo},
  author={Yao, Yao and Luo, Zixin and Li, Shiwei and Fang, Tian and Quan, Long},
  booktitle=ECCV,
  pages={767--783},
  year={2018}
}

@inproceedings{yao2019recurrent,
  title={Recurrent mvsnet for high-resolution multi-view stereo depth inference},
  author={Yao, Yao and Luo, Zixin and Li, Shiwei and Shen, Tianwei and Fang, Tian and Quan, Long},
  booktitle=CVPR,
  pages={5525--5534},
  year={2019}
}

@inproceedings{gu2020cascade,
  title={Cascade cost volume for high-resolution multi-view stereo and stereo matching},
  author={Gu, Xiaodong and Fan, Zhiwen and Zhu, Siyu and Dai, Zuozhuo and Tan, Feitong and Tan, Ping},
  booktitle=CVPR,
  pages={2495--2504},
  year={2020}
}

@inproceedings{cheng2020deep,
  title={Deep stereo using adaptive thin volume representation with uncertainty awareness},
  author={Cheng, Shuo and Xu, Zexiang and Zhu, Shilin and Li, Zhuwen and Li, Li Erran and Ramamoorthi, Ravi and Su, Hao},
  booktitle=CVPR,
  pages={2524--2534},
  year={2020}
}

@inproceedings{yang2020cost,
  title={Cost volume pyramid based depth inference for multi-view stereo},
  author={Yang, Jiayu and Mao, Wei and Alvarez, Jose M and Liu, Miaomiao},
  booktitle=CVPR,
  pages={4877--4886},
  year={2020}
}

@inproceedings{wang2021patchmatchnet,
  title={Patchmatchnet: Learned multi-view patchmatch stereo},
  author={Wang, Fangjinhua and Galliani, Silvano and Vogel, Christoph and Speciale, Pablo and Pollefeys, Marc},
  booktitle=CVPR,
  pages={14194--14203},
  year={2021}
}

@inproceedings{chen2019point,
  title={Point-based multi-view stereo network},
  author={Chen, Rui and Han, Songfang and Xu, Jing and Su, Hao},
  booktitle=ICCV,
  pages={1538--1547},
  year={2019}
}

@inproceedings{lin2021barf,
  title={Barf: Bundle-adjusting neural radiance fields},
  author={Lin, Chen-Hsuan and Ma, Wei-Chiu and Torralba, Antonio and Lucey, Simon},
  booktitle=ICCV,
  pages={5741--5751},
  year={2021}
}

@article{wang2021nerf,
  title={NeRF--: Neural radiance fields without known camera parameters},
  author={Wang, Zirui and Wu, Shangzhe and Xie, Weidi and Chen, Min and Prisacariu, Victor Adrian},
  year={2021}
}

@article{wu2023ifusion,
  title={ifusion: Inverting diffusion for pose-free reconstruction from sparse views},
  author={Wu, Chin-Hsuan and Chen, Yen-Chun and Solarte, Bolivar and Yuan, Lu and Sun, Min},
  journal={arXiv preprint arXiv:2312.17250},
  year={2023}
}

@inproceedings{xu2024sparp,
  title={Sparp: Fast 3d object reconstruction and pose estimation from sparse views},
  author={Xu, Chao and Li, Ang and Chen, Linghao and Liu, Yulin and Shi, Ruoxi and Su, Hao and Liu, Minghua},
  booktitle=ECCV,
  pages={143--163},
  year={2024},
  organization={Springer}
}

@article{lai2025hunyuan3d25,
  title={Hunyuan3D 2.5: Towards High-Fidelity 3D Assets Generation with Ultimate Details},
  author={Lai, Zeqiang and Zhao, Yunfei and Liu, Haolin and Zhao, Zibo and Lin, Qingxiang and Shi, Huiwen and Yang, Xianghui and Yang, Mingxin and Yang, Shuhui and Feng, Yifei and others},
  journal={arXiv preprint arXiv:2506.16504},
  year={2025}
}

@misc{chang2025reconviagenaccuratemultiview3d,
      title={ReconViaGen: Towards Accurate Multi-view 3D Object Reconstruction via Generation}, 
      author={Jiahao Chang and Chongjie Ye and Yushuang Wu and Yuantao Chen and Yidan Zhang and Zhongjin Luo and Chenghong Li and Yihao Zhi and Xiaoguang Han},
      year={2025},
      eprint={2510.23306},
      archivePrefix={arXiv},
      primaryClass={cs.CV},
      url={https://arxiv.org/abs/2510.23306}, 
}

@misc{hunyuan3d2025hunyuan3d21imageshighfidelity,
      title={Hunyuan3D 2.1: From Images to High-Fidelity 3D Assets with Production-Ready PBR Material}, 
      author={Team Hunyuan3D and Shuhui Yang and Mingxin Yang and Yifei Feng and Xin Huang and Sheng Zhang and Zebin He and Di Luo and Haolin Liu and Yunfei Zhao and Qingxiang Lin and Zeqiang Lai and Xianghui Yang and Huiwen Shi and Zibo Zhao and Bowen Zhang and Hongyu Yan and Lifu Wang and Sicong Liu and Jihong Zhang and Meng Chen and Liang Dong and Yiwen Jia and Yulin Cai and Jiaao Yu and Yixuan Tang and Dongyuan Guo and Junlin Yu and Hao Zhang and Zheng Ye and Peng He and Runzhou Wu and Shida Wei and Chao Zhang and Yonghao Tan and Yifu Sun and Lin Niu and Shirui Huang and Bojian Zheng and Shu Liu and Shilin Chen and Xiang Yuan and Xiaofeng Yang and Kai Liu and Jianchen Zhu and Peng Chen and Tian Liu and Di Wang and Yuhong Liu and Linus and Jie Jiang and Jingwei Huang and Chunchao Guo},
      year={2025},
      eprint={2506.15442},
      archivePrefix={arXiv},
      primaryClass={cs.CV},
      url={https://arxiv.org/abs/2506.15442}, 
}

@misc{hyper3d,
  title = {Hyper3D: High-Fidelity 3D Asset Generation},
  author = {{Hyper3D Team}},
  year = {2024},
  url = {https://hyper3d.ai/},
  note = {Accessed: 2024-05-20}
}

@misc{hitem3d,
  title = {Hitem3D: High-Quality 3D Model Generation Service},
  author = {{Hitem3D Team}},
  year = {2024},
  url = {https://www.hitem3d.ai/},
  note = {Accessed: 2024-05-20}
}

@misc{hy3d,
  title = {Hy-3D},
  author = {{Hy-3D Team}},
  year = {2024},
  url = {https://hy-3d.com},
  note = {Accessed: 2024-05-20}
}

@misc{tripo3d,
  title = {Tripo: Fast 3D Object Generation from Text and Image},
  author = {{Tripo AI}},
  year = {2024},
  url = {https://www.tripo3d.ai/},
  note = {Accessed: 2024-05-20}
}

@misc{xiang2025trellis2,
      title={Native and Compact Structured Latents for 3D Generation}, 
      author={Jianfeng Xiang and Xiaoxue Chen and Sicheng Xu and Ruicheng Wang and Zelong Lv and Yu Deng and Hongyuan Zhu and Yue Dong and Hao Zhao and Nicholas Jing Yuan and Jiaolong Yang},
      year={2025},
      eprint={2512.14692},
      archivePrefix={arXiv},
      primaryClass={cs.CV},
      url={https://arxiv.org/abs/2512.14692}, 
}

@inproceedings{Xue_2023_CVPR,
    author    = {Xue, Le and Gao, Mingfei and Xing, Chen and Mart{\'\i}n-Mart{\'\i}n, Roberto and Wu, Jiajun and Xiong, Caiming and Xu, Ran and Niebles, Juan Carlos and Savarese, Silvio},
    title     = {ULIP: Learning a Unified Representation of Language, Images, and Point Clouds for 3D Understanding},
    booktitle = {Proceedings of the IEEE/CVF Conference on Computer Vision and Pattern Recognition (CVPR)},
    month     = {June},
    year      = {2023},
    pages     = {1179-1189}
}

@inproceedings{zhou2023uni3d,
  title={Uni3d: Exploring unified 3d representation at scale},
  author={Zhou, Junsheng and Wang, Jinsheng and Ma, Baorui and Liu, Yu-Shen and Huang, Tiejun and Wang, Xinlong},
  booktitle={International Conference on Learning Representations (ICLR)},
  year={2024}
}

@article{chen2025ultra3d,
  title={Ultra3d: Efficient and high-fidelity 3d generation with part attention},
  author={Chen, Yiwen and Li, Zhihao and Wang, Yikai and Zhang, Hu and Li, Qin and Zhang, Chi and Lin, Guosheng},
  journal={arXiv preprint arXiv:2507.17745},
  year={2025}
}

@article{jang2016categorical,
  title={Categorical reparameterization with gumbel-softmax},
  author={Jang, Eric and Gu, Shixiang and Poole, Ben},
  journal={arXiv preprint arXiv:1611.01144},
  year={2016}
}

@article{lai2025lattice,
  title={LATTICE: Democratize High-Fidelity 3D Generation at Scale},
  author={Lai, Zeqiang and Zhao, Yunfei and Zhao, Zibo and Liu, Haolin and Lin, Qingxiang and Huang, Jingwei and Guo, Chunchao and Yue, Xiangyu},
  journal={arXiv preprint arXiv:2512.03052},
  year={2025}
}

@article{team2023gemini,
  title={Gemini: a family of highly capable multimodal models},
  author={Team, Gemini and Anil, Rohan and Borgeaud, Sebastian and Alayrac, Jean-Baptiste and Yu, Jiahui and Soricut, Radu and Schalkwyk, Johan and Dai, Andrew M and Hauth, Anja and Millican, Katie and others},
  journal={arXiv preprint arXiv:2312.11805},
  year={2023}
}

\clearpage
\appendix
\section*{Supplementary Materials}
\setcounter{section}{0}
\renewcommand{\thesection}{\Alph{section}}

\section{Quantitative Comparisons on ANYVIEW-200}
\label{sec:supp_anyview}

As discussed in the main paper (\S4.2), the \textsc{Anyview-200} benchmark comprises in-the-wild multi-view images from diverse sources without ground-truth 3D geometry. Consequently, geometry-based metrics such as Chamfer Distance and F-Score are not applicable. To provide a quantitative evaluation, we report the semantic similarity metrics ULIP-I~\cite{Xue_2023_CVPR} and Uni3D-I~\cite{zhou2023uni3d}, which measure the alignment between generated 3D shapes and input images in a shared embedding space.

\begin{table}[htbp]
  \caption{Quantitative comparison on \textsc{Anyview-200}. Since no ground-truth 3D geometry is available, we report semantic similarity metrics only. By default, metrics are computed by averaging the similarity between the generated 3D shape and all input views. For single-view methods, we additionally report a ``ref-only'' variant where the metric is computed against the reference view alone, providing a more favorable evaluation for methods that only observe a single image. Best results are in \textbf{bold}, second best are \underline{underlined}.}
  \label{tab:anyview200}
  \centering
  \setlength{\tabcolsep}{8pt}
  \begin{tabular}{l c c}
    \toprule
    Method & ULIP-I$\uparrow$ & Uni3D-I$\uparrow$ \\
    \midrule
    \multicolumn{3}{l}{\emph{Single-view generative models}} \\
    Hunyuan3D 2.0~\cite{hunyuan3d2025hunyuan3d21imageshighfidelity}       & 0.1372 & 0.3658 \\
    \quad + ref-only eval       & 0.1356 & 0.3737 \\
    Trellis2~\cite{xiang2025trellis2}       & 0.1391 & 0.3684 \\
    \quad + ref-only        & 0.1382 & 0.3790 \\
    \midrule
    \multicolumn{3}{l}{\emph{Multi-view reconstruction}} \\
    LGM~\cite{tang2025lgm}                  & 0.0969 & 0.2436 \\
    VGGT~\cite{wang2025vggt}                & 0.0858 & 0.2194 \\
    \midrule
    \multicolumn{3}{l}{\emph{Multi-view conditioned generation}} \\
    Hunyuan3D 2.0-MV~\cite{hunyuan3d2025hunyuan3d21imageshighfidelity}     & \underline{0.1404} & 0.3767 \\
    ReconViaGen~\cite{chang2025reconviagenaccuratemultiview3d} & 0.1343 & \underline{0.3806} \\
    \textbf{ROAR-3D (Ours) - stage 1}        & 0.1406 & 0.3859 \\
    \textbf{ROAR-3D (Ours) - stage 2}        & \textbf{0.1412} & \textbf{0.3916} \\
    \bottomrule
  \end{tabular}
\end{table}

In \cref{tab:anyview200}, we compare across all three categories. Single-view methods such as Trellis2 achieve reasonable scores but cannot leverage additional viewpoint information; even under a more favorable ref-only evaluation that considers only the reference view they observe, their scores remain below ours. Multi-view reconstruction methods (LGM, VGGT) yield the lowest scores, as they rely on geometric consistency assumptions that often fail on non-photographic, in-the-wild inputs. ReconViaGen achieves competitive results but remains sensitive to the quality of its internal reconstruction module. \methodName stage~1 already surpasses all baselines on both metrics, and stage~2 refinement further improves performance, confirming that our method generalizes well to unconstrained, diverse multi-view conditions.

\section{Scalability to Increasing Number of Input Views}
\label{sec:supp_viewcount}

As discussed in the main paper (\S4.1), \methodName is trained with 1 to 4 input views per instance. A natural question is whether the model generalizes to a larger number of views at inference time, particularly beyond the training range. We investigate this by varying the number of input views from 1 to 8 on a subset of objects from \textsc{Anyview-200}.

\begin{figure}[htbp]
  \centering
  \includegraphics[width=\linewidth]{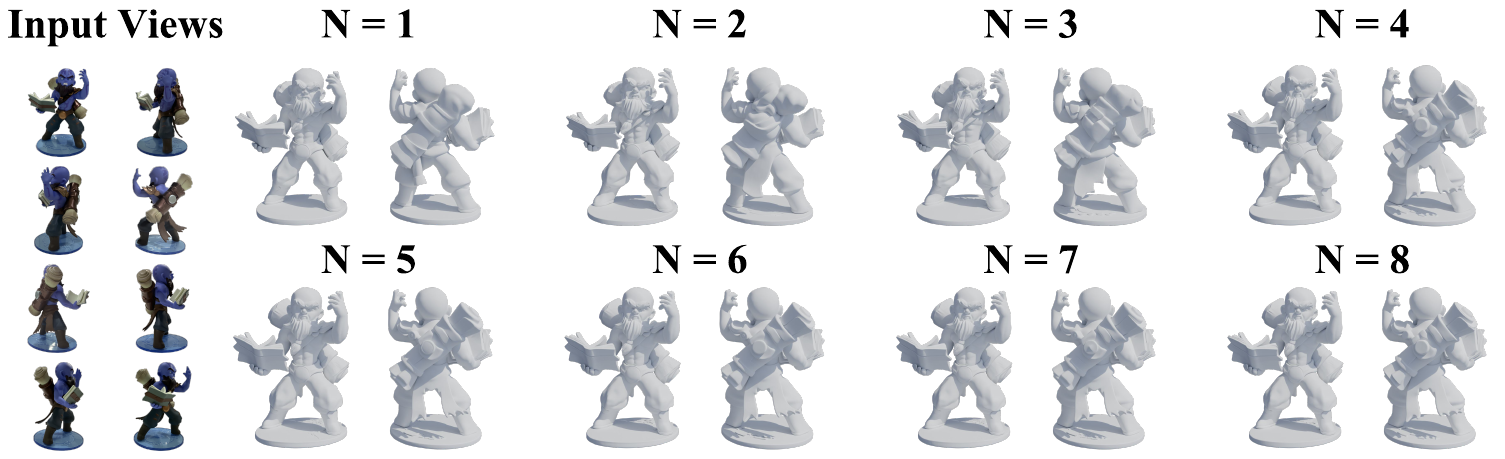}
  \caption{Effect of varying the number of input views. As more views are provided, the generated geometry becomes more complete and structurally accurate, particularly in regions initially unobserved. The model generalizes robustly beyond the training range (1--4 views) to 8 views.}
  \label{fig:view_scaling}
\end{figure}

As illustrated in \cref{fig:view_scaling}, we progressively increase the number of input views and examine the resulting 3D geometry. With only 1--2 frontal views, the model hallucinates plausible geometry for unobserved regions. As additional views covering new viewpoints are introduced, the generated geometry aligns more closely with the input images, particularly in previously unobserved areas. When further views from redundant viewpoints are added, the output remains stable with no degradation in quality. These results confirm that \methodName benefits monotonically from additional input views and generalizes reliably beyond its training-time view count, without requiring explicit camera pose estimation.

\section{Discussion on Partial-View and Close-Up Inputs}
\label{sec:supp_zoom}

\begin{figure}[htbp]
  \centering
  \includegraphics[width=\linewidth]{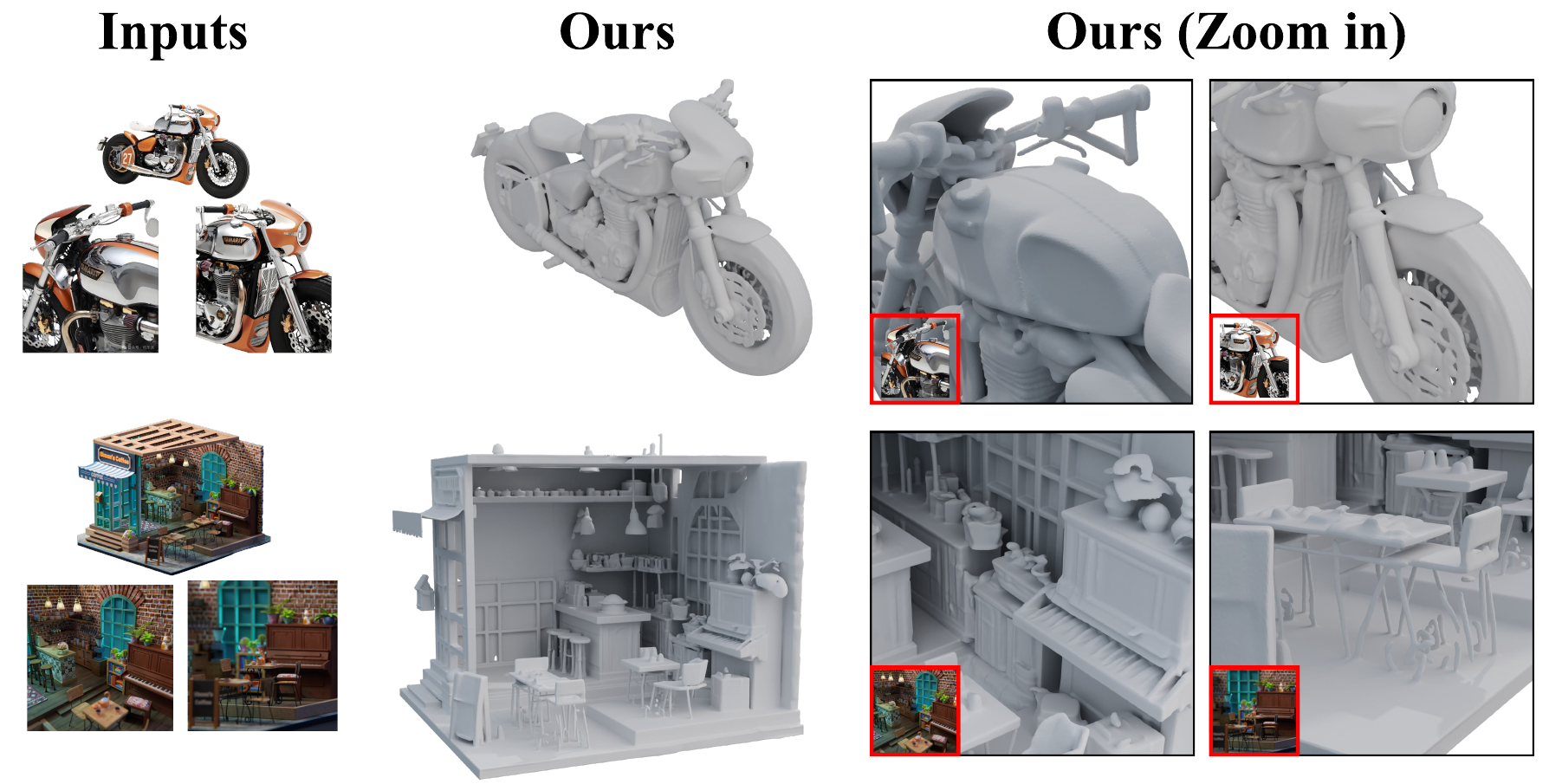}
  \caption{Effect of providing input views at different zoom levels. Close-up crops of specific regions enhance local geometric detail without affecting global shape.}
  \label{fig:zoom_levels}
\end{figure}

We investigate whether \methodName can handle inputs where some views are captured from close range, depicting only a local region of the object with different camera positions and orientations compared to the full-object view. As shown in \cref{fig:zoom_levels}, we provide a full-object image as the reference view together with close-range views of local regions as auxiliary views, which differ from the reference in distance, rotation, and translation simultaneously. Note that such partial-view inputs are entirely absent from our training data, and the results shown are from stage~1 only for illustration purposes. Despite this out-of-distribution setting, \methodName produces coherent 3D geometry that aligns well with both the global and local input views, without the close-range views causing artifacts or degrading the overall shape.

This robustness stems from the nature of our image-conditioned generation paradigm: by lifting visual information into 3D token space, the model implicitly establishes correspondences between arbitrary image pairs through 3D as a shared geometric bridge, regardless of the camera distance or viewing angle. This property opens up applications beyond arbitrary-view conditioned generation---for instance, leveraging the learned 3D representation for cross-image grounding or relative camera pose estimation between uncalibrated views---which we leave to future work.

\section{Ablation on Geometric Refinement}
\label{sec:supp_refinement}

\begin{figure}[htbp]
  \centering
  \includegraphics[width=\linewidth]{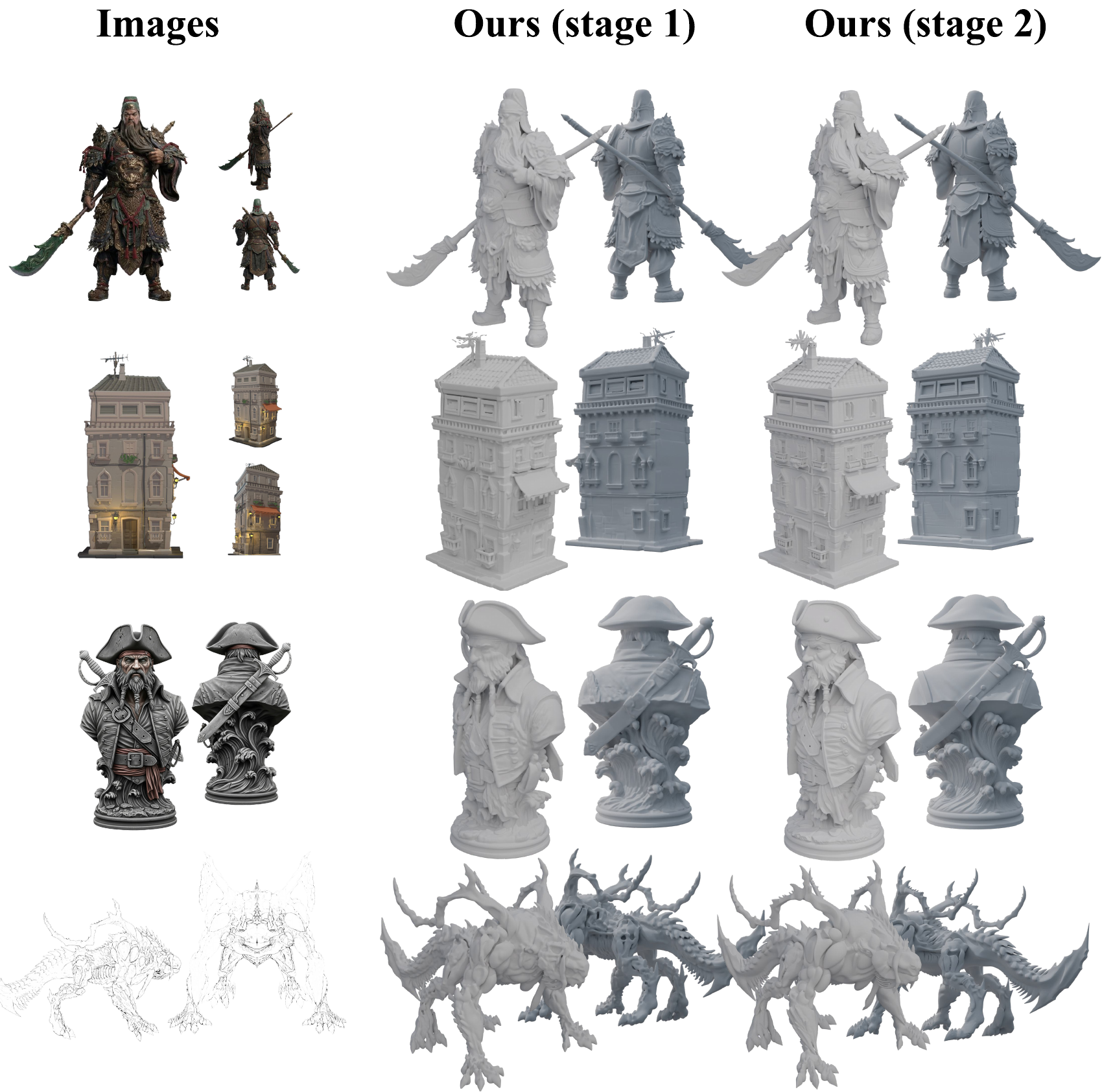}
  \caption{Ablation on geometric refinement. Stage~1 produces multi-view-consistent geometry, while stage~2 enhances fine-grained surface detail. Results span diverse input types including objects, architecture, human figures, and line-art sketches.}
  \label{fig:ablation_refine}
\end{figure}

As shown in \cref{fig:ablation_refine}, we visualize the outputs of both stages across diverse input types. Stage~1 already produces geometry that is well-aligned with the multi-view inputs, establishing correct global structure and proportions. Stage~2 then recovers higher-frequency geometric details such as facial features and architectural edges while preserving the overall structure from stage~1. Both stages perform multi-view-conditioned generation but at different levels of geometric granularity~\cite{lai2025lattice}: stage~1 captures the overall shape, and stage~2 complements it with fine-grained surface detail, as further supported by the quantitative gains reported in the main paper (\cref{tab:anyview200}).

% \section{Ablation on Geometric Refinement}
% \label{sec:supp_refinement}

% The proposed method consists of two stages: the first stage generates geometry from input images, and the second stage performs geometric refinement utilizing the results of the first stage. We further discuss the relationship between these two stages. 

% \begin{figure}[htbp]
%   \centering
%   \includegraphics[width=\linewidth]{img/sup_res_v1.pdf}
%   \caption{Ablation on geometric refinement. Stage~1 establishes multi-view-faithful geometry through the view router and dual-stream cross-attention, while stage~2 enhances fine-grained surface detail. Results span objects, architecture, human figures, and line-art sketches, demonstrating broad generalization.}
%   \label{fig:ablation_refine}
% \end{figure}

% As illustrated in \cref{fig:ablation_refine}, we present the input images alongside the 3D models generated in both stages. Facilitated by the multi-view capability introduced by the proposed view router and dual-stream cross-attention mechanisms, the 3D models obtained in the first stage already align closely with the multi-view input images. Subsequently, the second stage refines the models to achieve a cleaner and more detailed appearance, building upon the multi-view consistent structures established in the first stage. This observation further clarifies the division of labor between the two stages of the proposed method.

\section{Analysis on View Router}
\label{sec:supp_router}

To characterize the routing behavior of the proposed per-token view router, we measure consistency across 50 denoising timesteps and 21 transformer blocks on \textsc{Anyview-200}, as summarized in \cref{tab:router_consistency}.

The low cross-block consistency (0.4838) indicates that different transformer blocks attend to different views for the same token, distributing multi-view information across network depth. In contrast, cross-timestep consistency is substantially higher (0.7710), meaning each block's routing decision remains largely stable throughout the denoising process, with late blocks (14--20) reaching the highest agreement at 0.8121. The global consistency (0.4643), measured jointly across both blocks and timesteps, is close to the cross-block value, suggesting that cross-block variation is the dominant source of routing diversity while temporal stability is well preserved. Notably, the high cross-timestep consistency reveals considerable redundancy across denoising steps; exploiting this---\eg, by caching routing decisions across consecutive timesteps---could reduce computational cost, which we leave to future work.

\begin{table}[htbp]
\centering
\caption{Quantitative summary of router consistency metrics over 110 test samples.}
\label{tab:router_consistency}
\begin{tabular}{lcc}
\toprule
\textbf{Metric} & \textbf{Mean} & \textbf{Std} \\
\midrule
\textbf{Cross-Block Consistency} & 0.4838 & 0.0175 \\
\quad Early Steps (0--16) ($t \approx 1$) & 0.4745 & - \\
\quad Mid Steps (17--34) & 0.4905 & - \\
\quad Late Steps (35--49) ($t \approx 0$) & 0.4859 & - \\
\midrule
\textbf{Cross-Timestep Consistency} & 0.7710 & 0.1057 \\
\quad Early Blocks (0--6) & 0.7595 & - \\
\quad Mid Blocks (7--13) & 0.7412 & - \\
\quad Late Blocks (14--20) & 0.8121 & - \\
\midrule
\textbf{Global Consistency} & 0.4643 & 0.2321 \\
\bottomrule
\end{tabular}
\end{table}

\section{Limitations}
\label{sec:supp_limitations}

The proposed view router operates naturally within a cross-attention framework, where routing simply decides which view each 3D token attends to. Extending this to self-attention would require dense token-level attention masks whose cost grows quadratically with sequence length, making it less practical. We leave efficient self-attention-compatible routing to future work. In addition, the model has limited robustness to horizontally flipped or vertically inverted input images, since such augmentations were not included during training.

% \bibliography{main}

\end{document}